\documentclass{article}

\PassOptionsToPackage{numbers}{natbib}
\usepackage[final,main]{neurips_2026}
\makeatletter\renewcommand{\@noticestring}{Preprint}\makeatother
\makeatletter\renewcommand{\@bottomtitlebar}{\vskip 0.20in \vskip -\parskip \hrule height 1\p@ \vskip 0.07in}\makeatother
\usepackage[utf8]{inputenc}
\usepackage[T1]{fontenc}
\usepackage{hyperref}
\usepackage{url}
\usepackage{booktabs}
\usepackage{amsfonts}
\usepackage{amsmath}
\usepackage{nicefrac}
\usepackage{microtype}
\usepackage{xcolor}
\usepackage{colortbl}
\usepackage{graphicx}
\usepackage{multirow}
\usepackage{pifont}
\usepackage{todonotes}
\usepackage{soul}
\usepackage{enumitem}
\newcommand{\cmark}{\ding{51}}%
\newcommand{\xmark}{\ding{55}}%
\definecolor{oursrow}{RGB}{255,249,222}

\newcommand{\nickname}{ViDiHand}
\title{The Surprising Effectiveness of Video Diffusion Models for Hand Motion Reconstruction}

\author{
    Yuxi Wang$^{1}$\thanks{Equal contribution. \quad $^{\dagger}$Corresponding authors.}\hspace{0.35em},
    Chengkai Jin$^{1}$\footnotemark[1]\hspace{0.35em},
    Yufei Liu$^{2}$,
    Wenqi Ouyang$^{1}$,
    Tianyi Wei$^{1}$,\\
    \textbf{Zhiwei Zeng}$^{1}$,
    \textbf{Siyuan Huang}$^{2}$,
    \textbf{Zhiqi Shen}$^{1,\dagger}$,
    \textbf{Xingang Pan}$^{1,\dagger}$\\[2mm]
    $^{1}$Nanyang Technological University \qquad
    $^{2}$Shanghai Jiao Tong University
}

\hypersetup{bookmarksopen=true, bookmarksopenlevel=2}
\usepackage{float}

\begin{document}

\maketitle

\begin{figure}[h!]
  \centering
  \vspace{-0.65cm}
  \IfFileExists{figures/main_teaser.pdf}%
    {\includegraphics[width=\linewidth]{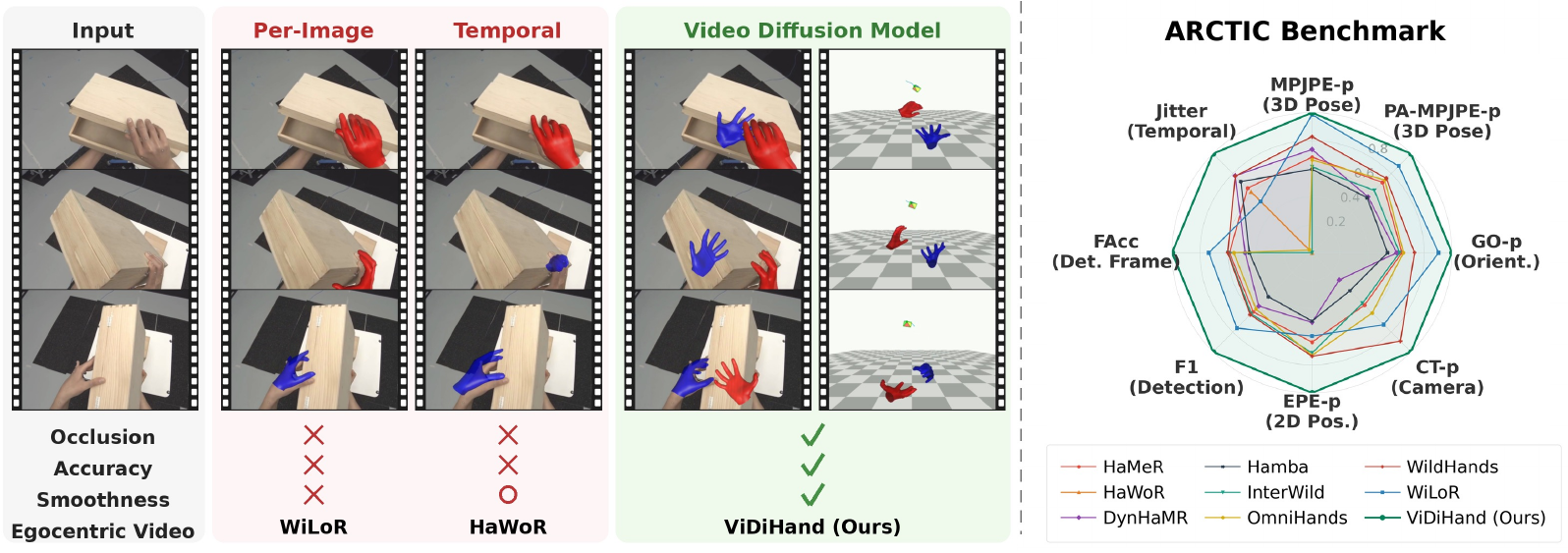}}%
    {\fbox{\parbox[c][3.0cm][c]{1.00\linewidth}{\centering\small Placeholder: \texttt{figures/main_teaser}.}}}
  \caption{\textbf{\nickname{} satisfies all three target properties of 4D hand recovery.} On the same egocentric input clip, the per-image baseline WiLoR~\cite{wilor_cvpr25} is sensitive to detection dropouts and suffers from frame-wise pose flicker; the temporal baseline OmniHands~\cite{omnihands_arxiv24} reduces flicker through cross-frame attention but still struggles under heavy occlusion and large hand-object motion. \nickname{} extracts features from a hand-aware video diffusion model and recovers a coherent and accurate 4D trajectory for both hands, with stable identity and smooth motion through occlusion.}
  \label{fig:teaser}
\end{figure}

\begin{abstract}

4D hand motion reconstruction from egocentric video is bottlenecked by two limitations: image-based pipelines depend on a detector that fails under heavy occlusion, while video-based methods rely on temporal modules learned only from scarce hand-pose annotations, too narrow to model motion, occlusion, and hand-object interaction. Yet these are exactly what video generative models must implicitly acquire to synthesize coherent video at internet scale. Motivated by this, we present \textit{\nickname{}}, which leverages the representations of a pretrained video diffusion model to reconstruct 4D two-hand pose. We adapt it via a hand-overlay rendering objective that specializes its features for hands while preserving its world priors. A decoder then recovers metric-scale pose from the adapted features. The pipeline operates directly on full frames—no detector, no infiller, no test-time optimization. On ARCTIC, HOT3D, and HOI4D, \nickname{} substantially outperforms prior methods on most metrics, establishing video diffusion models as a powerful new foundation for hand motion reconstruction and a route to scalable in-the-wild data collection for embodied AI.  Project page: \mbox{\url{https://vidihand.github.io}}.

\end{abstract}

\section{Introduction}
\label{sec:intro}

Reconstructing 4D hand motion from egocentric video is a persistent challenge, particularly given the pervasive occlusion in real-world human activity. The demand for solving it has surged with the rise of embodied AI, where egocentric video provides a natural and scalable source for robot learning. Recent efforts scale dexterous manipulation by pretraining on large amounts of egocentric human video~\cite{zheng2026egoscale,kareer2024egomimic,egodex,yang2025egovlalearningvisionlanguageactionmodels}, where hand/wrist trajectories serve as key supervision for imitation and policy learning. Thus, reconstructed hand-motion quality directly shapes policy learning from egocentric video at scale, where dense labels are scarce.

Yet existing hand recovery methods struggle with interaction-rich, occlusion-heavy video. Image-based methods~\cite{hamer_cvpr24,wilor_cvpr25,hamba_nips24,wildhands_eccv24,interwild_cvpr23}, built on image-pretrained backbones, depend on an upstream detector to reconstruct each frame independently. Under heavy occlusion, failed detections directly lead to reconstruction failure. Video-based methods attempt to address this in two ways, both relying on priors learned from scarce hand-labeled data. One line~\cite{haptic, fu2023deformerdynamicfusiontransformer, omnihands_arxiv24} adds cross-frame attention on top of an image-pretrained backbone, with only limited hand-pose supervision --- too narrow to learn the dynamics of motion, occlusion, and interaction from scratch. The other line~\cite{duran2023hmphandmotionpriors,dynhamr_cvpr25,hawor_cvpr25} introduces a learned motion prior or infiller trained on 3D hand trajectories alone, decoupled from the surrounding scene and the ongoing interaction, and thus still struggles with occlusion. These limitations call for representations beyond image-only and hand-only priors that instead \textit{capture the underlying geometry, motion, and interaction of the visual world in which the hands operate}.

Such representations are, in fact, increasingly available—in large-scale video generative models. Trained to synthesize temporally and geometrically coherent video at internet scale, video generative models must implicitly address the same structural challenges that 4D hand reconstruction faces: spatiotemporal consistency, 3D geometry from 2D observations, and reasoning about occluded content. The internal features of generative models have already been shown to support various vision tasks, including point tracking~\cite{ditracker}, dense prediction~\cite{visionbanana}, and 3D scene awareness~\cite{huang20253dvideofoundationmodels}. Yet, to our knowledge, no prior work has leveraged such rich priors for 4D hand reconstruction. 

We present \textbf{\nickname{}}, the first method to leverage a pretrained \textbf{Vi}deo \textbf{Di}ffusion model for 4D two-\textbf{Hand} motion reconstruction from egocentric video. Rather than treating the generative model as a frozen feature extractor, we adapt it through a hand-overlay rendering task that renders hand meshes onto the original frames. By editing only the hand region while reconstructing the rest of the scene, this adaptation steers the model's representation toward hand-aware reconstruction while preserving its prior knowledge. %
Built on these adapted features, a dual-branch decoder predicts both the relative 3D hand articulation and the per-joint 2D image-plane localization, whose combination anchors the hand at metric scale. The pipeline operates directly on full video frames, with no upstream hand detector, motion infiller, or test-time optimization.

\nickname{} outperforms prior methods, establishing a new state of the art on the most challenging hand reconstruction benchmarks. On the heavily occluded ARCTIC~\cite{arctic_cvpr23} sequences (Fig.~\ref{fig:teaser}), our method achieves near-perfect hand detection accuracy. Without any motion infiller or test-time optimization, \nickname{} still produces significantly smoother hand motion than all prior methods. On HOT3D's~\cite{hot3d_cvpr25} wide-angle fisheye footage and the cross-dataset HOI4D~\cite{hoi4d_cvpr22} benchmark, \nickname{} also leads on most metrics.
These gains are largely attributed to the video generative prior that brings internet-scale knowledge of occlusion, temporal coherence, and geometry to hand recovery for the first time. We view this as early evidence of a paradigm shift in 4D hand reconstruction: from hand-centric pipelines patched with specialized modules toward representations inherited from the increasingly powerful generative models the community is rapidly~advancing.

In summary, our contributions are:
\begin{itemize}[leftmargin=15pt,topsep=2pt]
\item \textbf{A new paradigm for hand motion reconstruction.} We present, to our knowledge, the first method to leverage a pretrained video diffusion model for 4D two-hand motion reconstruction from egocentric video. Our approach decodes high-quality hand motion from internal representations capturing rich structured information about the hands, manipulated objects, and scene. %
\item \textbf{Hand-overlay rendering as an adaptation target.} We identify hand-overlay rendering as an effective editing objective for adapting video diffusion models to hand reconstruction, specializing its features for hands while preserving its world priors. %
\item \textbf{State-of-the-art performance.} \nickname{} ranks first on nearly every metric: it cuts the detection error rate by up to $27\times$ and prediction jitter by at least $4\times$, and---under our coverage-aware protocol---reduces 2D end-point error by up to $4\times$ and leads 3D pose accuracy on all three benchmarks.
These results reveal the potential of video diffusion models for hand reconstruction and scalable data collection in embodied AI.

\end{itemize}

\section{Related Work}
\label{sec:related}

\paragraph{Monocular hand reconstruction.}
Monocular hand reconstruction recovers the 3D pose and shape of the hand from RGB input, commonly parameterized through the MANO~\cite{mano_sa17} model. Image-based methods estimate hand parameters from a single frame: HaMeR~\cite{hamer_cvpr24} demonstrated that a ViT-based regressor trained on heterogeneous 3D and 2D-keypoint datasets substantially improves generalization to in-the-wild data. WiLoR~\cite{wilor_cvpr25} extends this recipe with a coupled real-time detector--reconstructor trained on millions of images. Hamba~\cite{hamba_nips24} replaces dense ViT attention with graph-guided bidirectional Mamba scanning to improve robustness with fewer tokens. WildHands~\cite{wildhands_eccv24} addresses egocentric perspective distortion through intrinsics-aware positional encoding, and InterWild~\cite{interwild_cvpr23} targets two-hand interaction. Video-based methods attempt to address the temporal jitter, depth ambiguity, and detection failures left by per-frame estimators through various forms of temporal modeling. One line of work introduces cross-frame attention layers that fuse hand features across neighboring frames at the backbone level~\cite{haptic,omnihands_arxiv24,fu2023deformerdynamicfusiontransformer}. Another learns generative motion priors over 3D hand trajectories: HMP~\cite{duran2023hmphandmotionpriors} fits such a prior in test-time optimization, Dyn-HaMR~\cite{dynhamr_cvpr25} couples it with SLAM-based camera estimation, and HaWoR~\cite{hawor_cvpr25} applies it as a feedforward infiller. Across all of these, the reconstruction pipeline remains image-based or hand-centric, with no shared representation of the scene, object, and interaction context to disambiguate hand pose under heavy occlusion.

\paragraph{Video diffusion models and diffusion features.}
Video diffusion models have rapidly evolved from early designs such as the pixel-space VDM~\cite{vdm} and the latent-space SVD~\cite{svd} to large-scale text-to-video transformers including CogVideoX~\cite{cogvideox} and the Wan series~\cite{wan}, which now generate temporally and geometrically coherent video at billion-parameter, internet scale. A parallel line of work extends these models with controllable generation: VACE~\cite{vace} augments Wan2.1 with a unified conditioning path that supports inpainting, editing, and reference-based synthesis, while task-specific systems condition video generation on hand and body cues for embodied applications~\cite{wang2026hand2world, xie2026generatedrealityhumancentricworld}. Beyond generation, the internal representations of diffusion models have proven to be strong visual priors. On the image side, image diffusion models have been adapted to perception in several ways: Marigold~\cite{marigold} fine-tunes Stable Diffusion as a depth predictor, Vision Banana~\cite{visionbanana} reframes dense prediction as image generation, and DIFT~\cite{dift} extracts frozen diffusion features for semantic correspondence. Video diffusion features carry this further: a systematic comparison shows that the same architecture trained for video consistently outperforms its image-trained counterpart on spatial and motion-sensitive tasks~\cite{image2video}. Building on this, recent work analyzes how video diffusion transformers establish cross-frame correspondences in their attention layers and exploits them for zero-shot point tracking~\cite{yuan2026denoisetrackharnessingvideo,difftrack}, supervised tracking with diffusion-feature backbones~\cite{ditracker}, and 3D scene awareness~\cite{huang20253dvideofoundationmodels}. Yet no prior work has applied video diffusion priors to hand reconstruction.

\section{Method}
\label{sec:method}

\subsection{Overview}
\label{sec:overview}

Our central observation is that large-scale video diffusion models, trained to synthesize coherent video across diverse real-world scenes, must internally resolve the same three challenges that 4D hand recovery has long delegated to external modules: synthesizing content through occlusion, keeping stable identity and spatial placement across frames, and producing smooth motion. \nickname{} recovers per-frame MANO and camera translation for both hands directly from this internal representation, on full frames, with no detector, infiller, or test-time optimization.

\begin{figure}[t]
  \centering
  \IfFileExists{figures/main_pipeline.pdf}%
    {\includegraphics[width=1.00\linewidth]{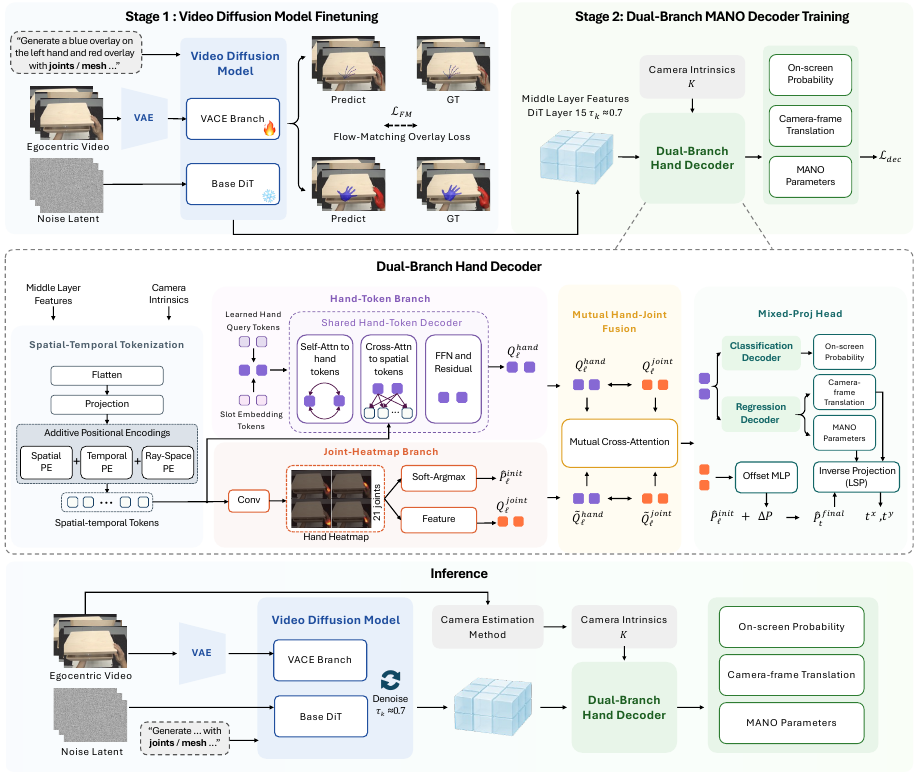}}%
    {\fbox{\parbox[c][3.5cm][c]{1.00\linewidth}{\centering\small Placeholder: \texttt{figures/main_pipeline.pdf}.}}}
  \caption{\textbf{\nickname{} pipeline.} \emph{Top:} the VACE branch is finetuned with hand-overlay rendering while the base DiT is frozen, producing a hand-aware video diffusion model. \emph{Middle:} the dual-branch decoder reads from a single $L^\star{=}15$, $\tau_k{\approx}0.7$ activation: a hand-token branch produces slot-aware summaries for articulated MANO pose, a parallel joint-heatmap branch produces 2D anchors and pooled descriptors for in-plane coordinates, mutual cross-attention couples them, and a mixed-projection head outputs MANO and depth while solving the in-plane translation in closed form against the heatmap. \emph{Bottom:} at inference, the same activation is decoded in one VACE pass.}
  \label{fig:pipeline}
\end{figure}

We achieve this in two stages. A \emph{hand-overlay rendering} pretext finetunes only the VACE branch of pretrained Wan2.1-VACE~\cite{vace,wan} to specialize the world prior to hands while preserving scene, object, and interaction priors (\S\ref{sec:vace-adaptation}). A \emph{dual-branch decoder} then recovers articulated MANO pose and image-space coordinates from the resulting representation, coupled by a closed-form geometric solve (\S\ref{sec:mano-decoder}). \S\ref{sec:losses} introduces the training objective; Fig.~\ref{fig:pipeline} summarizes the pipeline.

\subsection{Hand-Aware Video Diffusion Model}
\label{sec:vace-adaptation}

The pretrained Wan2.1-VACE backbone holds the priors we need, but they are entangled with everything else it knows about the visual world. The adaptation must therefore bind the representation to MANO hand geometry while leaving its scene, object, and motion priors intact. 

\paragraph{Hand-overlay rendering.}
We supervise the VACE branch alone, keeping the base DiT frozen, to regenerate the input clip with a semi-transparent rendered hand overlay alpha-blended onto each hand at every frame, including frames where the hand is fully occluded by an object. The objective is the standard flow-matching loss; no MANO-parameter supervision is applied at this stage. We use a two-stage curriculum: 
Stage~1a renders 2D joint-skeleton overlays, exposing the model to egocentric hand--object motion at a scale unavailable in MANO-annotated data; Stage~1b switches to MANO mesh overlays to align the representation to the MANO surface the decoder consumes. 
Forcing coherent overlays through occluded frames pushes the backbone to maintain a per-hand 3D state rather than merely inpaint the visible pixels; only the VACE branch is updated, keeping the base DiT and its world prior untouched (Tab.~\ref{tab:data-ablation}).

\paragraph{Feature extraction.}
The pose-relevant signal is not evenly distributed across DiT layers and denoising steps. Among the $30$ transformer blocks of the 1.3B-parameter Wan2.1-VACE backbone, the mid-block activation at layer $L^\star{=}15$ preserves enough spatial resolution to localize joints and enough abstraction to commit to articulated pose. 
Along the denoising axis, we capture features at $\tau_k{\approx}0.7$, where $\tau_k$ denotes the normalized denoising step, running from $\tau{=}0$ (pure noise) to $\tau{=}1$ (fully denoised); at this step the latent has not yet committed to the rendered overlay appearance. The captured feature $\mathbf{F}=\{F_\ell\}_{\ell=1}^{N_{\mathrm{lat}}}$ consists of $N_{\mathrm{lat}}{=}21$ per-latent-frame activations $F_\ell$ on the DiT spatial grid for each 81-frame clip. The decoder of \S\ref{sec:mano-decoder} is trained on $\mathbf{F}$ alone, without backpropagating through the diffusion backbone; the choice $(L^\star,\tau_k)$ is validated empirically (Tabs.~\ref{tab:layer-ablation}, \ref{tab:timestep-ablation}).

\subsection{Dual-Branch Hand Decoder}
\label{sec:mano-decoder}

The hand-aware backbone holds the per-hand 3D state implicitly inside its activations, and surfacing it requires a decoder that respects two structurally different axes of the representation. Articulated MANO pose is a holistic property of the hand: any single joint angle is determined only when the rest of the hand is also accounted for. Image-space joint coordinates, in contrast, are local: each joint sits at a position on the spatial token grid that decouples from the others. A single regressor compressing both into one token would conflate two structurally different kinds of evidence. We therefore use two parallel branches, a hand-token branch specialized for articulated pose and a joint-heatmap branch specialized for image coordinates, coupled by one mutual cross-attention layer; a mixed-projection head splits camera translation into a regressed depth and a closed-form in-plane shift (Fig.~\ref{fig:pipeline}, middle).

\paragraph{Spatio-temporal tokenization.}
Each feature $F_\ell$ is projected to decoder width $h$ by a learned matrix $W_F$ and augmented with three additive positional codes, yielding spatio-temporal tokens $X_\ell$,
\begin{equation}
X_\ell = \mathrm{LN}(F_\ell W_F^\top) + P^{\mathrm{sp}} + P^{\mathrm{tmp}}_\ell + g_{\mathrm{ray}}\!\big(\Gamma(K)\big),
\label{eq:tokenizer}
\end{equation}
where $\mathrm{LN}$ is layer normalization, $P^{\mathrm{sp}}$ and $P^{\mathrm{tmp}}_\ell$ are learned spatial/temporal embeddings, $\Gamma(K)$ is a sinusoidal encoding of camera-ray azimuth and elevation from the per-clip intrinsics $K$, and $g_{\mathrm{ray}}$ is a small MLP lifting it to token width. The ray-space embedding lets a single decoder serve heterogeneous camera intrinsics without per-dataset specialization.

\paragraph{Hand-token branch.}
Two learned slot queries cross-attend to $X_\ell$ through stacked transformer-decoder layers, producing $Q^{\mathrm{hand}}_\ell \in \mathbb{R}^{2\times h}$; slot identity is fixed by the query row, removing the need for a handedness classifier. Each query integrates evidence across the entire hand to commit to a full 3D configuration, matching the holistic structure of articulated pose.

\paragraph{Joint-heatmap branch.}
A $1{\times}1$ convolutional head maps the tokens $X_\ell$ to per-joint logit maps $\mathcal{H}_\ell$, which a spatial softmax normalizes to per-joint attention weights $A_\ell$; these read out a 2D anchor $\widehat{\mathbf P}^{\mathrm{init}}_\ell$ from the token grid $\mathbf{x}^{\mathrm{grid}}$ and a pooled visual descriptor $Q^{\mathrm{joint}}_\ell$ from the tokens:
\begin{equation}
A_\ell = \mathrm{softmax}(\mathcal{H}_\ell),\qquad
\widehat{\mathbf P}^{\mathrm{init}}_\ell = A_\ell\,\mathbf{x}^{\mathrm{grid}},\qquad
Q^{\mathrm{joint}}_\ell = A_\ell\,X_\ell.
\label{eq:heatmap-branch}
\end{equation}
Each joint coordinate is anchored directly to the spatial token grid along which the diffusion backbone organizes its visual content, matching the local, per-joint nature of image-space coordinates.

\paragraph{Mutual hand--joint fusion.}
The two branches carry complementary evidence: hand tokens encode the articulated configuration of each hand, while joint descriptors localize each joint in the image. One mutual cross-attention layer exchanges this information,
\begin{equation}
\begin{aligned}
\widetilde Q^{\mathrm{hand}}_\ell  &= \mathrm{LN}\big(Q^{\mathrm{hand}}_\ell + \mathrm{MHA}(Q^{\mathrm{hand}}_\ell,Q^{\mathrm{joint}}_\ell,Q^{\mathrm{joint}}_\ell)\big),\\
\widetilde Q^{\mathrm{joint}}_\ell &= \mathrm{LN}\big(Q^{\mathrm{joint}}_\ell + \mathrm{MHA}(Q^{\mathrm{joint}}_\ell,Q^{\mathrm{hand}}_\ell,Q^{\mathrm{hand}}_\ell)\big),
\end{aligned}
\label{eq:xattn}
\end{equation}
where $\mathrm{MHA}(q,k,v)$ is multi-head attention; $\widetilde Q^{\mathrm{hand}}$ now carries joint-level image evidence (consumed by the MANO regressor) and $\widetilde Q^{\mathrm{joint}}$ articulated-pose context (refining the 2D anchors).

\paragraph{Mixed-projection head.}
Camera translation has two components with structurally different image evidence: depth $t^z$ is monocularly ambiguous and admits no closed-form solution from a single view, whereas the in-plane shift $(t^x,t^y)$ is well-conditioned given depth, joints, and intrinsics. We therefore split the translation, regressing only the depth and solving the in-plane shift in closed form. After temporal upsampling of the latent-frame tokens to video frames $t$, an MLP $h_{\mathrm{MANO}}$ regresses MANO orientation, pose, shape, log-depth, and an on-screen probability:
\begin{equation}
(\widehat{\mathbf R}_t,\,\widehat{\boldsymbol\Theta}_t,\,\widehat{\mathbf B}_t,\,\widehat{\boldsymbol\zeta}_t,\,\widehat{\mathbf e}_t)=h_{\mathrm{MANO}}(\widetilde Q^{\mathrm{hand}}_t),\qquad \widehat{\mathbf t}^z_t=\exp(\widehat{\boldsymbol\zeta}_t),
\label{eq:mano-head}
\end{equation}
and an offset MLP $h_{\mathrm{off}}$ refines the heatmap anchor to $\widehat{\mathbf P}^{\mathrm{final}}_t = \widehat{\mathbf P}^{\mathrm{init}}_t + h_{\mathrm{off}}(\widetilde Q^{\mathrm{joint}}_t)$. With canonical joints $\widehat{\mathbf J}^{\mathrm{can}}_t = \mathcal{M}(\widehat{\mathbf R}_t,\widehat{\boldsymbol\Theta}_t,\widehat{\mathbf B}_t)$ from the differentiable MANO forward $\mathcal{M}$ and depth $\widehat{\mathbf t}^z_t$ fixed, the pinhole projection under intrinsics $K{=}(f_x,f_y,c_x,c_y)$
\begin{equation}
\hat u_j = f_x\,\frac{X_j^{\mathrm{can}} + t^x}{Z_j^{\mathrm{can}} + \widehat{\mathbf t}^z_t} + c_x,
\qquad
\hat v_j = f_y\,\frac{Y_j^{\mathrm{can}} + t^y}{Z_j^{\mathrm{can}} + \widehat{\mathbf t}^z_t} + c_y
\label{eq:pinhole}
\end{equation}
is linear in the unknown in-plane shift $(t^x,t^y)$ and decouples per coordinate. The shift is therefore recovered as a per-coordinate weighted least-squares fit of the projection to the refined anchors $\widehat{\mathbf P}^{\mathrm{final}}_t$, weighted by the anchor-visibility mask $M^{\mathrm{dir}}$ (closed form in the supplementary); at inference the solve uses only predicted quantities. Because the solve is differentiable, the heatmap, offset MLP, MANO regressor, and depth scalar are jointly optimized as one geometric system, avoiding the root-translation/root-pose ambiguity that arises when $(t^x,t^y)$ is regressed freely alongside MANO. The camera-frame joints follow as $\widehat{\mathbf J}_t = \widehat{\mathbf J}^{\mathrm{can}}_t + \widehat{\mathbf t}_t$.

\subsection{Training Objective}
\label{sec:losses}

The decoder is trained with a sum of five terms,
\begin{equation}
\mathcal{L}_{\mathrm{dec}}=\mathcal{L}_{\mathrm{MANO}}+\mathcal{L}_{\mathrm{cam}}+\mathcal{L}_{\mathrm{img}}+\mathcal{L}_{\mathrm{vis}}+\mathcal{L}_{\mathrm{temp}},
\label{eq:loss}
\end{equation}
each playing a distinct role in the coupled MANO--camera system. $\mathcal{L}_{\mathrm{MANO}}$ supervises global orientation and articulated rotations with geodesic losses on $\mathrm{SO}(3)$ and shape with MSE. $\mathcal{L}_{\mathrm{cam}}$ supervises the assembled translation and camera-frame joints, anchoring the closed-form solve. $\mathcal{L}_{\mathrm{img}}$ couples MANO and camera through one image residual, jointly supervising the refined 2D anchors that drive the in-plane solve and the pinhole-projected MANO joints so the two branches agree on the same pixel positions. $\mathcal{L}_{\mathrm{vis}}$ is a BCE on the on-screen probability over all slots (including empty ones), suppressing hallucinated hands. $\mathcal{L}_{\mathrm{temp}}$ adds a translation-acceleration $\ell_1$ term and a stop-gradient shape-consistency term, applied only at training time. More details are in the supplementary.

\section{Experiments}
\label{sec:experiments}

\subsection{Experimental Setup}
\label{sec:datasets}

\paragraph{Datasets.}
We evaluate on three egocentric hand benchmarks chosen to stress complementary failure modes. \textbf{ARCTIC}~\cite{arctic_cvpr23} concentrates severe hand--object and hand--hand occlusion under bimanual manipulation of articulated objects. \textbf{HOT3D}~\cite{hot3d_cvpr25} pairs a wide-angle fisheye lens with high-dynamic-range lighting and rapid head/hand motion, stressing detection under distortion, motion blur, and bright--dark co-exposure; since its official test set lacks ground-truth MANO, we randomly hold out 5\% of validation sequences as our test split. \textbf{HOI4D}~\cite{hoi4d_cvpr22} is held out from every method's explicit supervised training and used for evaluation only; ``held out'' refers to explicit supervision, since the video backbone's undisclosed pretraining corpus cannot be audited (\S\ref{sec:conclusion}). For training, \textbf{EgoDex}~\cite{egodex} contributes joint-only egocentric supervision for adapting the video model. Each segment is an 81-frame clip; full per-dataset specifications and preprocessing details are in the supplementary.

\paragraph{Evaluation protocol.}
Standard per-hand metrics such as MPJPE and PA-MPJPE only score correctly matched true-positive predictions, biasing evaluation toward methods that conservatively skip hard frames, precisely the frames on which occlusion-aware methods should be tested. We therefore adopt a \emph{coverage-aware evaluation protocol} (metrics carry a ``-p'' suffix): each of the $n_{\mathrm{TP}}$ true positives contributes its error $e_i$, and each of the $n_{\mathrm{FN}}$ false negatives is charged the placeholder error $e_i^{\mathrm{can}}$ of an identity-rotation, mean-shape MANO placed at the camera origin,
\begin{equation}
  \mathrm{metric}_{\text{-}\mathrm{p}} = \frac{\sum_{i\in\mathrm{TP}} e_i + \sum_{i\in\mathrm{FN}} e_i^{\mathrm{can}}}
  {n_{\mathrm{TP}} + n_{\mathrm{FN}}}\,.
\end{equation}
The placeholder error is computed per sample from the missed ground-truth hand, so the false-negative cost is deterministic and method-independent. Per-metric placeholder definitions, and a comparison on co-predicted hands (scoring only hands both methods detect), are in the supplementary.

We report nine metrics. \emph{Detection}: recall, F1, and frame accuracy FAcc, where FAcc is the fraction of frames in which all on-screen ground-truth hands are correctly matched with no hallucinated hand. \emph{3D pose}: MPJPE-p and PA-MPJPE-p, both in mm. \emph{Orientation and position}: 2D end-point error EPE-p in px, geodesic global-orientation error GO-p in degrees, and camera-translation error CT-p in m. \emph{Temporal}: jitter in mm/frame$^2$. Detection metrics are higher-better; the rest are lower-better.

\subsection{Comparison with State of the Art}
\label{sec:sota}

\paragraph{Baselines.}
We compare against eight methods: four single-image regressors (HaMeR~\cite{hamer_cvpr24}, WiLoR~\cite{wilor_cvpr25}, Hamba~\cite{hamba_nips24}, InterWild~\cite{interwild_cvpr23}); two with egocentric or two-hand specialization (WildHands~\cite{wildhands_eccv24}, OmniHands~\cite{omnihands_arxiv24}); and two world-frame video methods (Dyn-HaMR~\cite{dynhamr_cvpr25} with test-time SLAM-guided optimization, HaWoR~\cite{hawor_cvpr25} with adaptive egocentric SLAM and a motion infiller). All baselines are evaluated on the same 81-frame segments under the protocol of \S\ref{sec:datasets}. Tab.~\ref{tab:main-comparison} reports results on in-distribution ARCTIC and HOT3D and held-out HOI4D.

\begin{table}[t]
\caption{\textbf{Comparison on three egocentric benchmarks.} \emph{ARCTIC}~\cite{arctic_cvpr23} and \emph{HOT3D}~\cite{hot3d_cvpr25} are in-distribution; \emph{HOI4D}~\cite{hoi4d_cvpr22} is held out from explicit supervised training for all methods. All 3D-pose and orientation/position columns use the coverage-aware protocol of \S\ref{sec:datasets}, folding detection coverage into every pose metric. \textbf{Bold}: best per column within each block.}
\label{tab:main-comparison}
\centering
\resizebox{\textwidth}{!}{%
\begin{tabular}{@{}c@{\hspace{8pt}} l ccc cc ccc c@{}}
\toprule
& \multirow{2}{*}{\textbf{Method}} & \multicolumn{3}{c}{\textbf{Detection}} & \multicolumn{2}{c}{\textbf{3D Pose}} & \multicolumn{3}{c}{\textbf{Orient.\ \& Position}} & \textbf{Temporal} \\
\cmidrule(lr){3-5} \cmidrule(lr){6-7} \cmidrule(lr){8-10} \cmidrule(lr){11-11}
& & FAcc & Recall & F1 & MPJPE-p & PA-MPJPE-p & EPE-p & GO-p & CT-p & Jitter \\
\midrule
\multirow{9}{*}{\rotatebox[origin=c]{90}{\textbf{ARCTIC}}}
  & InterWild~\cite{interwild_cvpr23}   & 0.878 & 0.943 & 0.959 & 30.817 & 15.952 & 53.888 & 25.386 & 0.097 & 46.577 \\
  & HaMeR~\cite{hamer_cvpr24}           & 0.875 & 0.943 & 0.957 & 29.197 & 14.596 & 65.289 & 24.907 & 0.095 & 18.279 \\
  & Hamba~\cite{hamba_nips24}           & 0.833 & 0.912 & 0.941 & 31.233 & 17.168 & 87.047 & 27.822 & 0.110 & 15.357 \\
  & WildHands~\cite{wildhands_eccv24}   & 0.879 & 0.946 & 0.960 & 25.704 & 13.941 & 50.517 & 22.320 & 0.058 & 12.972 \\
  & OmniHands~\cite{omnihands_arxiv24}  & 0.866 & 0.949 & 0.954 & 29.674 & 14.203 & 51.505 & 24.580 & 0.087 & 45.312 \\
  & WiLoR~\cite{wilor_cvpr25}           & 0.919 & 0.951 & 0.974 & 22.012 & 11.873 & 71.527 & 17.358 & 0.075 & 24.091 \\
  & Dyn-HaMR~\cite{dynhamr_cvpr25}      & 0.842 & 0.918 & 0.951 & 27.904 & 17.017 & 85.723 & 25.951 & 0.121 & 12.840 \\
  & HaWoR~\cite{hawor_cvpr25}           & 0.700 & 0.817 & 0.895 & 45.357 & 26.375 & 158.062 & 43.325 & 0.149 & 19.789 \\
\rowcolor{oursrow}
  & \textbf{\nickname{} (Ours)} & \textbf{0.997} & \textbf{0.999} & \textbf{0.999} & \textbf{21.668} & \textbf{9.821} & \textbf{12.407} & \textbf{14.642} & \textbf{0.047} & \textbf{3.183} \\
\midrule
\multirow{9}{*}{\rotatebox[origin=c]{90}{\textbf{HOT3D}}}
  & InterWild~\cite{interwild_cvpr23}   & 0.669 & 0.881 & 0.868 & 77.168 & 24.811 & 71.482 & 58.501 & 0.213 & 101.164 \\
  & HaMeR~\cite{hamer_cvpr24}           & 0.692 & 0.904 & 0.883 & 68.314 & 21.455 & 59.077 & 49.636 & 0.102 & 23.632 \\
  & Hamba~\cite{hamba_nips24}           & 0.632 & 0.828 & 0.853 & 71.732 & 29.620 & 107.625 & 56.525 & 0.128 & 18.507 \\
  & WildHands~\cite{wildhands_eccv24}   & 0.655 & 0.863 & 0.844 & 52.791 & 28.946 & 111.438 & 53.933 & 0.157 & 22.885 \\
  & OmniHands~\cite{omnihands_arxiv24}  & 0.649 & 0.895 & 0.868 & 63.281 & 22.682 & 68.437 & 49.120 & 0.133 & 69.510 \\
  & WiLoR~\cite{wilor_cvpr25}           & 0.827 & 0.897 & 0.937 & 30.966 & 19.980 & 72.978 & 25.746 & 0.098 & 17.976 \\
  & Dyn-HaMR~\cite{dynhamr_cvpr25}      & 0.614 & 0.811 & 0.802 & 74.214 & 38.201 & 171.617 & 43.851 & 0.571 & 44.942 \\
  & HaWoR~\cite{hawor_cvpr25}           & 0.348 & 0.499 & 0.654 & 71.396 & 66.031 & 327.294 & 79.350 & 0.262 & 23.872 \\
\rowcolor{oursrow}
  & \textbf{\nickname{} (Ours)} & \textbf{0.948} & \textbf{0.974} & \textbf{0.983} & \textbf{21.514} & \textbf{11.383} & \textbf{14.953} & \textbf{15.829} & \textbf{0.040} & \textbf{3.741} \\
\midrule
\multirow{9}{*}{\rotatebox[origin=c]{90}{\textbf{HOI4D}}}
  & InterWild~\cite{interwild_cvpr23}   & 0.731 & 0.922 & 0.864 & 53.072 & 22.909 & 80.549 & 41.743 & 0.228 & 98.866 \\
  & HaMeR~\cite{hamer_cvpr24}           & 0.731 & 0.923 & 0.864 & 44.481 & 21.580 & 79.494 & 33.557 & 0.187 & 20.068 \\
  & Hamba~\cite{hamba_nips24}           & 0.710 & 0.885 & 0.849 & 47.161 & 25.924 & 115.793 & 37.390 & 0.204 & 21.556 \\
  & WildHands~\cite{wildhands_eccv24}   & 0.730 & 0.924 & 0.864 & 45.623 & 23.601 & 82.246 & 45.654 & 0.159 & 18.615 \\
  & OmniHands~\cite{omnihands_arxiv24}  & 0.655 & 0.937 & 0.834 & 44.255 & 18.689 & 70.662 & 34.392 & \textbf{0.108} & 24.212 \\
  & WiLoR~\cite{wilor_cvpr25}           & 0.962 & 0.966 & 0.972 & 33.710 & 14.903 & 41.579 & 25.527 & 0.115 & 17.449 \\
  & Dyn-HaMR~\cite{dynhamr_cvpr25}      & 0.750 & 0.863 & 0.845 & 45.097 & 29.259 & 144.643 & 40.176 & 0.258 & 17.947 \\
  & HaWoR~\cite{hawor_cvpr25}           & 0.869 & 0.864 & 0.919 & 47.329 & 28.851 & 135.748 & 43.091 & 0.139 & 28.376 \\
\rowcolor{oursrow}
  & \textbf{\nickname{} (Ours)} & \textbf{0.984} & \textbf{0.991} & \textbf{0.990} & \textbf{30.090} & \textbf{13.960} & \textbf{24.460} & \textbf{23.420} & 0.117 & \textbf{4.010} \\
\bottomrule
\end{tabular}%
}
\end{table}

\paragraph{Quantitative results.}
Our method ranks first on $26$ of $27$ metrics across the three benchmarks; we analyze the gains along three axes: detection robustness, pose accuracy, and temporal smoothness.
For \emph{detection robustness}, ARCTIC frame accuracy reaches $0.997$, up from $0.919$ for the strongest discriminative baseline WiLoR, cutting the frame-error rate $27\times$ (from $8.1\%$ to $0.3\%$). FAcc requires \emph{every} on-screen hand to be correctly recovered, the regime where detector-driven baselines drop one of two interacting hands. 

For \emph{pose accuracy}, \nickname{} attains the best MPJPE-p and PA-MPJPE-p on all three benchmarks. The margin is widest on HOT3D, where MPJPE-p improves from the best baseline's $31.0$\,mm to $21.5$\,mm; on ARCTIC, EPE-p falls from $50.5$\,px (WildHands) to $12.4$\,px, a $4\times$ reduction. 
In the supplementary, we further report metrics on co-predicted hands, which isolates per-hand accuracy from detection coverage. Under this setting, \nickname{} still outperforms most baselines across the majority of metrics.

For \emph{smoothness}, prediction jitter on ARCTIC drops to $3.18$, $4\times$ below the smoothest prior method, Dyn-HaMR's $12.8$. Dyn-HaMR reaches that figure \emph{with} SLAM-guided test-time optimization, whereas \nickname{} uses neither a motion prior nor test-time optimization. Smoothness thus reflects backbone coherence plus a lightweight training-time regularizer, not inference-time engineering. The lead carries over to held-out HOI4D, where \nickname{} ranks first on eight of nine metrics, cutting jitter to $4.0$ (vs.\ $17.4$ for the next-best method) and EPE-p to $24.5$ from $41.6$.

\subsection{Ablation Studies}
\label{sec:ablations}

We ablate the DiT layer (Tab.~\ref{tab:layer-ablation}), the denoising step (Tab.~\ref{tab:timestep-ablation}), the backbone adaptation (Tab.~\ref{tab:data-ablation}), and the decoder components (Tab.~\ref{tab:decoder-ablation}). To isolate each axis, all ablations restrict the decoder training to ARCTIC alone and evaluate on the ARCTIC test set. A controlled fitting study and loss-component ablations are in the supplementary.

\begin{table}[t]
\centering
\begin{minipage}[t]{0.46\textwidth}
\centering
\caption{\textbf{Layer ablation.} DiT feature-layer sweep at $\tau_k{\approx}0.7$. Selected: $L_{15}$.}
\label{tab:layer-ablation}
\resizebox{0.88\textwidth}{!}{%
\begin{tabular}{lcccc}
\toprule
Layer & FAcc & MPJPE-p & EPE-p & Jitter \\
\midrule
$L_{8}$            & 0.9972 & 22.38 & 13.60 & 3.71 \\
$\mathbf{L_{15}}$  & \textbf{0.9979} & \textbf{20.59} & \textbf{11.93} & \textbf{3.42} \\
$L_{22}$           & 0.9978 & 22.51 & 12.85 & 3.46 \\
$L_{29}$           & 0.9972 & 24.46 & 14.29 & 3.73 \\
\bottomrule
\end{tabular}
}
\end{minipage}
\hfill
\begin{minipage}[t]{0.51\textwidth}
\centering
\caption{\textbf{Denoising step ablation.} Normalized denoising step $\tau_k$ sweep at $L_{15}$ ($\tau{=}0$ pure noise, $\tau{=}1$ fully denoised). Selected: $\tau_k{\approx}0.7$.}
\label{tab:timestep-ablation}
\resizebox{0.9\textwidth}{!}{%
\begin{tabular}{lcccc}
\toprule
Denoising step & FAcc & MPJPE-p & EPE-p & Jitter \\
\midrule
$\tau_k{\approx}0.3$               & 0.9967 & 20.93 & 11.95 & \textbf{3.39} \\
$\tau_k{\approx}0.5$               & 0.9970 & 20.73 & 12.13 & \textbf{3.39} \\
$\boldsymbol{\tau_k{\approx}0.7}$  & \textbf{0.9979} & \textbf{20.59} & \textbf{11.93} & 3.42 \\
$\tau_k{\approx}0.9$               & 0.9974 & 22.17 & 13.27 & 3.41 \\
\bottomrule
\end{tabular}%
}
\end{minipage}
\end{table}

\paragraph{Feature layer and denoising step.}
The hand-pose signal localizes to mid-block features read out partway along the denoising trajectory. Along the layer axis (Tab.~\ref{tab:layer-ablation}), $L_{15}$, the 15th of $30$ DiT blocks, performs best on every metric: the early $L_{8}$ is still tied to low-level pixel features, while the late $L_{22}/L_{29}$ are already biased toward the rendered overlay texture. Along the denoising axis (Tab.~\ref{tab:timestep-ablation}), $\tau_k{\approx}0.7$ attains the best FAcc, MPJPE-p, and EPE-p. 
VACE’s editing formulation keeps the reference video as control at every step, so the backbone reads scene structure even when the latent is mostly noise. 
As $\tau_k$ approaches the fully-denoised end, the latent resolves into the rendered overlay appearance. The drop at $\tau_k{\approx}0.9$ thus reflects features committed to the rendered mesh rather than the underlying interaction. Jitter is essentially flat across both sweeps.

\begin{table}[t]
\centering
\begin{minipage}[t]{0.46\textwidth}
\centering
\caption{\textbf{Backbone adaptation ablation.} ``Mesh overlay'' uses Stage~1b; ``Joint+Mesh'' adds Stage~1a. Selected: Joint+Mesh.}
\label{tab:data-ablation}
\resizebox{\textwidth}{!}{%
\begin{tabular}{lcccc}
\toprule
Backbone / adaptation & FAcc & MPJPE-p & EPE-p & Jitter \\
\midrule
Randomly initialized & 0.9795 & 36.39 & 32.94 & 14.68 \\
DINOv3  & 0.9852 & 24.46 & 18.93 & 14.14 \\
Pretrained T2V       & 0.9932 & 21.64 & 15.15 & \textbf{3.14} \\
Pretrained VACE      & 0.9846 & 22.67 & 16.60 & 3.47 \\
Mesh overlay         & 0.9965 & 21.23 & 12.75 & 3.20 \\
Joint+Mesh overlay   & \textbf{0.9979} & \textbf{20.59} & \textbf{11.93} & 3.42 \\
\bottomrule
\end{tabular}%
}
\end{minipage}
\hfill
\begin{minipage}[t]{0.51\textwidth}
\centering
\caption{\textbf{Decoder-component ablation.} Each row removes one component from the full decoder while keeping the feature slice and training protocol fixed.}
\label{tab:decoder-ablation}
\resizebox{\textwidth}{!}{%
\begin{tabular}{lcccc}
\toprule
Variant & FAcc & MPJPE-p & EPE-p & Jitter \\
\midrule
w/o ray-space PE              & 0.9829 & 21.52 & 12.36 & \textbf{2.84} \\
w/o Joint-Heatmap Branch      & 0.9819 & 22.20 & 14.91 & 3.41 \\
w/o Hand--Joint Fusion        & 0.9809 & 20.64 & 13.82 & 3.01 \\
w/o Mixed-Projection solve    & 0.9767 & \textbf{20.26} & 16.26 & 2.86 \\
\midrule
Full decoder                  & \textbf{0.9979} & 20.59 & \textbf{11.93} & 3.42 \\
\bottomrule
\end{tabular}%
}
\end{minipage}
\end{table}

\paragraph{Backbone adaptation.}
We fix the decoder and vary only how the video backbone is adapted (Tab.~\ref{tab:data-ablation}). 
A transformer trained from random initialization lags the pretrained backbones across all metrics. DINOv3-H+ (840M)~\cite{simeoni2025dinov3}, a large-scale image backbone, also underperforms our 1.3B Wan backbone. The largest gap is in jitter, suggesting video pretraining yields more temporally stable representations. The pretrained T2V and un-adapted VACE backbones lack the spatial precision for joint localization (EPE-p $15.15$ and $16.60$); Stage~1b mesh-overlay supervision lowers EPE-p to $12.75$, and adding Stage~1a joint-overlay further reduces it to $11.93$ by exposing the model to egocentric hand--object motion unavailable in MANO-annotated data. 

\paragraph{Decoder components.}
Every component contributes to detection: removing any one drops FAcc from $0.9979$ to $0.9767$--$0.9829$ (Tab.~\ref{tab:decoder-ablation}). The Joint-Heatmap Branch and the mixed-projection solve dominate 2D localization: replacing the former with direct 2D-coordinate regression raises EPE-p by $2.98$\,px and MPJPE-p by $1.61$\,mm, and disabling the latter raises EPE-p by $4.33$\,px, confirming that heatmap localization and the geometric in-plane solve drive image-space placement. Disabling the solve also slightly lowers MPJPE-p and jitter ($20.59$ to $20.26$ and $3.42$ to $2.86$), a trade-off we accept for its FAcc and EPE-p gains. Removing the ray-space PE or the Hand--Joint Fusion lowers jitter slightly but degrades detection, MPJPE-p, and EPE-p. We adopt the full configuration for its detection and 2D-placement lead, not for topping every column.

\subsection{Qualitative Results}
\label{sec:robustness}

\begin{figure}[t]
  \centering
  \IfFileExists{figures/main_qualitative.pdf}%
    {\includegraphics[width=1.00\linewidth]{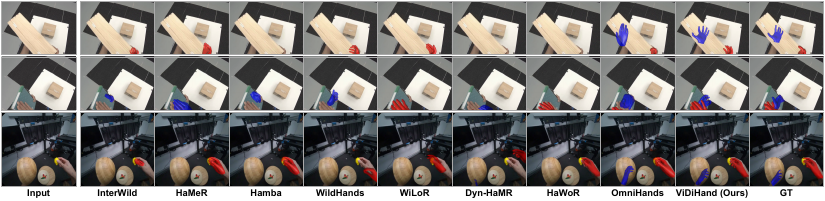}}%
    {\fbox{\parbox[c][3.0cm][c]{1.00\linewidth}{\centering\small Placeholder: \texttt{figures/main_qualitative.pdf}.}}}
  \caption{\textbf{Qualitative comparison on ARCTIC and HOT3D under severe occlusion.} \textbf{Top:} one hand fully occluded behind a box; most baselines drop it or swap its handedness. \textbf{Middle:} one hand is self-occluded near the frame boundary; only \nickname{} recovers it. \textbf{Bottom:} one hand severely occluded by a bowl; most baselines output no mesh, while \nickname{} matches GT throughout.}
  \label{fig:qualitative}
\end{figure}

\begin{figure}[t]
  \centering
  \IfFileExists{figures/main_qualitative_in_the_wild.pdf}%
    {\includegraphics[width=1.00\linewidth]{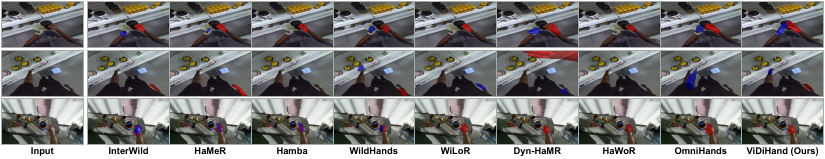}}%
    {\fbox{\parbox[c][3.0cm][c]{1.00\linewidth}{\centering\small Placeholder: \texttt{figures/main_qualitative_in_the_wild.pdf}.}}}
  \caption{\textbf{Qualitative comparison on in-the-wild data.} \textbf{Top:} severe occlusion by a towel and a jar, several baselines miss one or both hands. \textbf{Middle:} one hand reaches into a shelf, no baselines generate plausible result. \textbf{Bottom:} single-hand scene with grating-like shadows, multiple baselines hallucinate a second hand. \nickname{} produces plausible reconstructions across all three cases.}
  \label{fig:qualitative_in_the_wild}
\end{figure}

Figs.~\ref{fig:qualitative} and~\ref{fig:qualitative_in_the_wild} compare \nickname{} with baselines on challenging cases. In all cases, baselines drop the occluded hand, misalign its articulation, or hallucinate a second hand, whereas \nickname{} recovers both hands with plausible articulation. These are precisely the occlusion and truncation regimes in which the adapted backbone maintains a per-hand 3D state that detector-driven pipelines cannot. Additional comparisons are in the supplementary.

\section{Conclusion}
\label{sec:conclusion}

We presented \nickname{}, which recovers 4D two-hand motion from egocentric video using a pretrained video diffusion prior. A hand-overlay objective specializes the backbone onto the MANO surface while preserving its scene, object, and motion priors; a dual-branch decoder then recovers articulated pose and image-space coordinates from the adapted representation, with camera translation anchored by a closed-form mixed-projection solve.

\nickname{} ranks first on every coverage-aware metric on ARCTIC and HOT3D and on eight of nine on held-out HOI4D, without inference-time smoothing, pairing near-complete detection coverage with competitive per-hand accuracy. As backbones scale, recovering structured 3D state from their internal representations offers a path to scalable in-the-wild 4D hand annotation.

\paragraph{Limitations and future work.}
At $5.5$\,fps on $4$ A100 GPUs, \nickname{} is currently an offline annotation tool; distillation and few-step or autoregressive generators are the most direct path toward real-time inference. Training still relies on pose-labeled video, but full MANO is not essential: joint-only annotations already drive the Stage-1a overlay pretext. Scaling supervision through weaker labels, image datasets, and additional camera viewpoints, and extending beyond hand--object interaction to full-body motion, is therefore a natural next step.

\paragraph{Acknowledgements.}
This research is supported by RIE2025 Industry Alignment Fund – Industry Collaboration Projects (IAF-ICP) (Award I2301E0026), administered by A*STAR, and by Alibaba Group and NTU Singapore through the Alibaba-NTU Global e-Sustainability CorpLab (ANGEL).

\bibliographystyle{plain}
\bibliography{refs}

\clearpage

\appendix
\renewcommand{\thesection}{\Alph{section}}

\begin{center}
\Large\textbf{Supplementary Material}
\end{center}
\vspace{0.5em}

\noindent
This supplement is organized so that the formal evaluation protocol comes first and the results that depend on it follow. \S\ref{sec:supp-eval} derives the coverage-aware evaluation protocol introduced in the main paper and gives the closed-form definition of every evaluation metric. \S\S\ref{sec:supp-tponly} and~\ref{sec:supp-detection} extend the main-paper comparison along two axes: a pairwise pose comparison on co-predicted hands across all three benchmarks, and a per-side breakdown of detection F1. \S\ref{sec:supp-qualitative} adds thirteen additional qualitative comparison figures drawn from ARCTIC, HOT3D, and in-the-wild egocentric video. \S\ref{sec:supp-implementation} specifies the datasets, decoder architecture, loss functions, and two-stage training procedure. \S\ref{sec:supp-ablations} reports a loss-term ablation on ARCTIC.

\vspace{0.5em}
\tableofcontents
\vspace{1em}

\section{Evaluation Protocol and Metric Definitions}
\label{sec:supp-eval}

We report nine metrics organized in four categories (Tab.~\ref{tab:metric-summary}). We first define each metric (\S\ref{sec:supp-metrics}), then describe the prediction--ground-truth alignment procedure (\S\ref{sec:supp-alignment}), and finally introduce the \emph{coverage-aware evaluation protocol} (\S\ref{sec:supp-penalty}) that folds false negatives into every pose metric so that detection coverage and pose accuracy are captured by a single number. Metrics computed under it carry a ``-p'' subscript (for \emph{presence}: every present ground-truth hand contributes).

\begin{table}[!htbp]
\caption{\textbf{Evaluation metrics.} Category, unit, and whether the coverage-aware protocol of \S\ref{sec:supp-penalty} applies. Detection metrics are higher-better; the rest are lower-better. All pose metrics use the coverage-aware protocol; Jitter is the only metric that opts out, since the second-order finite difference requires three contiguous tracked observations.}
\label{tab:metric-summary}
\centering\small
\begin{tabular}{llcc}
\toprule
Category & Metric & Unit & Coverage-aware? \\
\midrule
\multirow{3}{*}{Detection}
  & FAcc                & -- & -- \\
  & Recall              & -- & -- \\
  & F1                  & -- & -- \\
\midrule
\multirow{2}{*}{3D Pose}
  & MPJPE-p             & mm & \cmark \\
  & PA-MPJPE-p          & mm & \cmark \\
\midrule
\multirow{3}{*}{Orient.\,\&\,Pos.}
  & EPE-p               & px & \cmark \\
  & GO-p                & $^\circ$ & \cmark \\
  & CT-p                & m  & \cmark \\
\midrule
Temporal
  & Jitter              & mm/frame$^2$ & \xmark \\
\bottomrule
\end{tabular}
\end{table}

\subsection{Metric Definitions}
\label{sec:supp-metrics}

We now define each of the nine metrics. Throughout, we use $\widehat{\mathbf{J}}_j \in \mathbb{R}^3$ for the $j$-th predicted camera-frame joint position, with index $j \in \{0, 1, \ldots, J{-}1\}$ and $J = 21$; $j{=}0$ is the wrist and $j = 1, \ldots, J{-}1$ are the remaining MANO joints in OpenPose order. The same 21-joint ordering is used by every loss term in \S\ref{sec:supp-losses} and every evaluation metric in this section. We use $\mathbf{J}^\star_j$ for the corresponding ground truth.

\paragraph{Detection metrics.}
The three detection metrics characterize how reliably a method detects and correctly classifies hands by side (left/right), independently of pose accuracy. They are computed on corpus-level counts aggregated across all frames and segments in the test set.

\emph{Frame accuracy (FAcc).}
Frame accuracy is the strictest detection measure: it is the fraction of frames in which every on-screen ground-truth hand is correctly matched \emph{and} no on-screen false-positive prediction exists. A single on-screen missed hand, a single on-screen extra prediction, or a left--right swap on any side causes the entire frame to fail:
\begin{equation}
\mathrm{FAcc} = \frac{\bigl|\{t : \mathrm{FP}^{\mathrm{os}}_t = 0 \;\wedge\; \mathrm{FN}^{\mathrm{os}}_t = 0\}\bigr|}{|\mathcal{F}|}\,,
\label{eq:fa}
\end{equation}
where $\mathcal{F}$ is the set of all frames containing at least one ground-truth hand, and $\mathrm{FP}^{\mathrm{os}}_t$, $\mathrm{FN}^{\mathrm{os}}_t$ are the on-screen false-positive and false-negative counts in frame $t$ under the off-screen-exclusion policy of \S\ref{sec:supp-alignment}. Off-screen ground-truth hands and predictions matched to off-screen hands are excluded from this count, so a frame whose only errors involve hands entirely outside the field of view still counts as perfect; this convention is used consistently throughout the paper, namely that methods are not penalized for failing to predict invisible hands. Frame accuracy is particularly informative for downstream tasks that require reliable two-hand input on every frame (e.g., robot policy learning), because a single missed frame can corrupt a trajectory.

\emph{Recall.}
Recall is the fraction of ground-truth hand instances (across all frames) that are successfully detected and matched to a prediction of the correct handedness:
\begin{equation}
\mathrm{Recall} = \frac{n_{\mathrm{TP}}}{n_{\mathrm{TP}} + n_{\mathrm{FN}}}\,.
\label{eq:recall}
\end{equation}
Recall counts are aggregated per side (left, right) and then summed. A method with high recall but low precision tends to produce spurious extra hands, whereas a method with high precision but low recall misses hands in difficult frames.

\emph{F1 score.}
F1 is the harmonic mean of precision and recall, balancing false-positive and false-negative rates:
\begin{equation}
\mathrm{Precision} = \frac{n_{\mathrm{TP}}}{n_{\mathrm{TP}} + n_{\mathrm{FP}}}\,,
\qquad
\mathrm{F1} = \frac{2 \cdot \mathrm{Precision} \cdot \mathrm{Recall}}{\mathrm{Precision} + \mathrm{Recall}}\,.
\label{eq:f1}
\end{equation}
Precision penalizes spurious predictions (predictions with no corresponding ground truth), while recall penalizes missed detections, so F1 summarizes both failure modes in a single number.

\paragraph{3D pose metrics (coverage-aware).}
The 3D pose metrics measure the accuracy of the predicted hand skeleton. Both operate on root-relative joint positions to factor out absolute translation.

\emph{MPJPE-p (mm).}
MPJPE-p is Mean Per-Joint Position Error after root alignment. For each true-positive matched pair, the 21 predicted joints are root-aligned by subtracting the wrist (joint~0) position, and compared to the similarly root-aligned ground truth. The per-sample error is
\begin{equation}
e_{\mathrm{MPJPE}} = \frac{1}{J} \sum_{j=0}^{J-1} \bigl\|(\widehat{\mathbf{J}}_j - \widehat{\mathbf{J}}_0) - (\mathbf{J}^\star_j - \mathbf{J}^\star_0)\bigr\|_2\,,
\label{eq:mpjpe}
\end{equation}
where $J = 21$ and all positions are in meters; the final metric is reported in millimeters ($\times 1000$). Root alignment removes the effect of absolute camera-frame translation, isolating the error in the hand's internal articulation and global orientation. For false negatives, the canonical MANO placeholder (Eq.~\ref{eq:canonical}) produces a per-sample error of about $132$\,mm on ARCTIC, reflecting the displacement between the rest pose and a typical articulated ground-truth hand. The coverage-aware metric is then computed via Eq.~\ref{eq:penalty}.

\emph{PA-MPJPE-p (mm).}
PA-MPJPE-p is Procrustes-aligned MPJPE. Before computing the per-joint distance, we stack the root-relative predicted joints $\widehat{\mathbf{P}} = \{\widehat{\mathbf{J}}_j - \widehat{\mathbf{J}}_0\}_{j=0}^{J-1} \in \mathbb{R}^{J \times 3}$ and the root-relative ground truth $\mathbf{P}^\star = \{\mathbf{J}^\star_j - \mathbf{J}^\star_0\}_{j=0}^{J-1} \in \mathbb{R}^{J \times 3}$, and align $\widehat{\mathbf{P}}$ to $\mathbf{P}^\star$ via a similarity transformation (rotation, translation, and uniform scale) that minimizes the sum of squared distances. Concretely, we first center both point clouds, $\overline{\widehat{\mathbf{P}}} = \widehat{\mathbf{P}} - \boldsymbol{\mu}_{\hat{P}}$ and $\overline{\mathbf{P}}^\star = \mathbf{P}^\star - \boldsymbol{\mu}_{P^\star}$, where $\boldsymbol{\mu}$ denotes the column-wise mean. We then take the SVD of the cross-covariance matrix
\begin{equation}
\mathbf{H} = \overline{\widehat{\mathbf{P}}}^{\!\top} \overline{\mathbf{P}}^\star = \mathbf{U} \boldsymbol{\Sigma} \mathbf{V}^{\!\top},
\label{eq:procrustes-svd}
\end{equation}
recover the optimal rotation while handling reflections,
\begin{equation}
\mathbf{R}^{\mathrm{opt}} = \mathbf{V}\,\mathrm{diag}\!\bigl(1,\; 1,\; \det(\mathbf{V}\mathbf{U}^{\!\top})\bigr)\,\mathbf{U}^{\!\top},
\label{eq:procrustes-rot}
\end{equation}
and the optimal scale,
\begin{equation}
s^{\mathrm{opt}} = \frac{\mathrm{tr}\!\bigl(\mathbf{R}^{\mathrm{opt}}\mathbf{H}\bigr)}{\mathrm{tr}\!\bigl(\overline{\widehat{\mathbf{P}}}^{\!\top}\overline{\widehat{\mathbf{P}}}\bigr)}\,.
\label{eq:procrustes-scale}
\end{equation}
The aligned prediction and the resulting per-sample error are
\begin{equation}
\widehat{\mathbf{P}}^{\mathrm{aligned}} = s^{\mathrm{opt}}\,\overline{\widehat{\mathbf{P}}}\,(\mathbf{R}^{\mathrm{opt}})^{\!\top} + \boldsymbol{\mu}_{P^\star}\,,
\qquad
e_{\mathrm{PA}} = \frac{1}{J}\sum_{j=0}^{J-1} \bigl\|\widehat{\mathbf{P}}^{\mathrm{aligned}}_{j} - \mathbf{P}^{\star}_{j}\bigr\|_2\,,
\label{eq:pa-mpjpe}
\end{equation}
where $\widehat{\mathbf{P}}^{\mathrm{aligned}}_{j}$ and $\mathbf{P}^\star_j$ denote the $j$-th row of the corresponding matrix. On matched true positives, Procrustes alignment removes global orientation, translation, and scale, so this term isolates \emph{articulation} error: two matched predictions with equal Procrustes-aligned error but different MPJPE differ in global wrist rotation and hand scale, not in finger articulation. As a coverage-aware metric, PA-MPJPE-p additionally charges the raw canonical placeholder (below) for every missed hand, so the reported value also reflects detection coverage and is not a pure articulation score.

For false negatives, the canonical placeholder uses the \emph{raw} (non-Procrustes-aligned) canonical MPJPE of about $132$\,mm, consistent with the MPJPE-p placeholder. Applying Procrustes alignment to the canonical rest pose would reduce the FN cost to roughly $10$--$25$\,mm (because the alignment has enough degrees of freedom to absorb much of the displacement), which would make missed detections nearly costless and undermine the coverage-aware protocol.

\paragraph{Orientation and position metrics (coverage-aware).}
The orientation and position metrics assess the predicted global wrist orientation, absolute 3D position in camera frame, and 2D reprojection accuracy. Unlike the 3D pose metrics above, they are \emph{not} root-relative and therefore capture absolute spatial accuracy.

\emph{GO-p (${}^\circ$).}
GO-p is the geodesic global-orientation error. The per-sample error is the geodesic (shortest-path) distance between the predicted and ground-truth wrist rotation matrices on $\mathrm{SO}(3)$, which is geometrically the angle of the unique axis-angle rotation that maps one orientation to the other:
\begin{equation}
e_{\mathrm{GO}} = \arccos\!\Bigl(\mathrm{clamp}\!\bigl(\tfrac{\mathrm{tr}(\hat R^{\!\top} R^\star) - 1}{2},\; {-1},\; 1\bigr)\Bigr) \cdot \frac{180}{\pi}\,,
\label{eq:go}
\end{equation}
where $\hat R, R^\star \in \mathrm{SO}(3)$ are the predicted and ground-truth global-orientation (wrist) rotation matrices, and the $\mathrm{clamp}$ operation ensures numerical stability when the trace is near the boundary values $-1$ or $3$. The result is in degrees. For false negatives, the placeholder is the geodesic distance from the identity matrix to the ground-truth orientation: $e_{\mathrm{GO}}^{\mathrm{can}} = \mathrm{geodesic}(\mathbf{I}_3,\, R^\star)$. The placeholder varies per sample, ranging from about $40^\circ$ to $90^\circ$ on ARCTIC, and reflects the actual wrist rotation of the missed hand.

\emph{CT-p (m).}
CT-p is the camera-translation error. The per-sample error is the Euclidean distance between the predicted and ground-truth camera-frame translations (3D wrist positions):
\begin{equation}
e_{\mathrm{CT}} = \bigl\|\widehat{\mathbf{t}} - \mathbf{t}^\star\bigr\|_2\,,
\label{eq:ct}
\end{equation}
where $\widehat{\mathbf{t}}, \mathbf{t}^\star \in \mathbb{R}^3$ are the predicted and ground-truth camera translation vectors in meters. CT-p is the only metric that measures absolute 3D wrist placement in camera frame; the MPJPE family cancels translation via root alignment. Camera translation includes depth ($t^z$) and is coupled to focal length, making it sensitive to intrinsics modeling. For false negatives, the placeholder is the distance from the camera origin to the ground-truth hand, $e_{\mathrm{CT}}^{\mathrm{can}} = \|\mathbf{t}^\star\|_2$; its magnitude is dataset-dependent, roughly $0.5$--$0.8$\,m on chest-mounted ARCTIC, slightly shorter on head-mounted HOT3D (Aria), and up to about $1.5$\,m on HOI4D where the camera can swing far from the hand.

\emph{EPE-p (px).}
EPE-p is the 2D end-point error. Each 3D joint is projected onto the image plane via pinhole projection with the segment's intrinsic parameters; depth is floored at $z \geq 0.01$\,m before division to prevent pinhole blow-up. Unlike the mm/deg/m coverage-aware metrics, EPE-p aggregates at the \emph{per-joint} level rather than the per-sample level, with a per-joint on-screen mask: only joints whose \emph{ground-truth} 2D projection lies in $[0, W) \times [0, H)$ with $z > 0.01$\,m contribute to either the numerator or the denominator. Concretely,
\begin{equation}
\text{EPE-p} = \frac{
  \sum_{(i,j)\in\mathcal{T}}\min\!\Bigl(\bigl\|\pi_K(\widehat{\mathbf{J}}_{i,j}) - \pi_K(\mathbf{J}^\star_{i,j})\bigr\|_2,\; d_{\mathrm{img}}\Bigr)
  \;+\; d_{\mathrm{img}}\cdot|\mathcal{N}|}
{|\mathcal{T}| + |\mathcal{N}|}\,,
\label{eq:epe}
\end{equation}
where $\mathcal{T} = \{(i,j) : \mathrm{TP}\ i,\ \mathrm{joint}\ j\ \text{on-screen}\}$ is the set of true-positive (sample, joint) pairs and $\mathcal{N} = \{(i,j) : \mathrm{FN}\ i,\ \text{ground-truth joint}\ j\ \text{on-screen}\}$ is the corresponding set of false-negative (sample, joint) pairs. The pinhole projection is
\begin{equation}
\pi_K(\mathbf{p}) = \begin{pmatrix} f_x \, p_x / p_z + c_x \\ f_y \, p_y / p_z + c_y \end{pmatrix}\,,
\label{eq:supp-pinhole}
\end{equation}
and $d_{\mathrm{img}} = \sqrt{W^2 + H^2}$ is the image diagonal. Per-joint distances are clamped to $d_{\mathrm{img}}$ to prevent numerical blow-up when $p_z \approx 0$ pushes a pinhole projection to extreme pixel coordinates. For each false-negative ground-truth hand, every on-screen joint contributes a placeholder distance of $d_{\mathrm{img}}$, giving roughly $826$\,px on ARCTIC ($672 \times 480$), $679$\,px on HOT3D ($480 \times 480$), and $980$\,px on HOI4D ($854 \times 480$). The image-diagonal choice is the worst-case pixel distance any in-frame joint pair can attain, giving a bounded but substantial per-joint pixel cost for a missed hand, larger than the roughly $200$\,px image-center distance a midpoint placeholder would give.

\paragraph{Temporal metric.}

\emph{Jitter (mm/frame$^2$).}
Jitter measures the temporal smoothness of predicted 3D joint trajectories via the mean magnitude of the second-order finite difference (discrete acceleration). It is computed on \emph{contiguous runs}: a run is a maximal sequence of consecutive frames in which the same ground-truth hand identity is continuously matched to a prediction. If a frame has a false negative for a given hand side, the run on that side is broken. Only runs with $L_r \geq 3$ frames are included, since the second difference requires at least three consecutive observations.

For a single run $r$ of length $L_r$, let $\widehat{\mathbf{J}}_{j,t}^{(r)}$ denote the absolute camera-frame 3D position of joint $j$ in video frame $t$. The per-run jitter is
\begin{equation}
\mathrm{Jitter}_r = \frac{1}{(L_r - 2) \cdot J} \sum_{t=2}^{L_r - 1} \sum_{j=0}^{J-1} \bigl\|\widehat{\mathbf{J}}_{j,t+1}^{(r)} - 2\,\widehat{\mathbf{J}}_{j,t}^{(r)} + \widehat{\mathbf{J}}_{j,t-1}^{(r)}\bigr\|_2\,,
\label{eq:jitter-run}
\end{equation}
and the global jitter is a weighted average across all runs, where each run contributes proportionally to its number of acceleration samples:
\begin{equation}
\mathrm{Jitter} = \frac{\displaystyle\sum_{r} (L_r - 2) \cdot \mathrm{Jitter}_r}{\displaystyle\sum_{r} (L_r - 2)}\,.
\label{eq:jitter-global}
\end{equation}
The result is in mm/frame$^2$ (positions are converted from meters to millimeters), and lower values indicate smoother, more temporally coherent predictions. Jitter captures the temporal instability of predicted poses without reference to ground-truth acceleration: a method can have low jitter but high MPJPE-p (smooth but wrong), or the reverse. We compute Jitter on true positives only (no coverage-aware placeholder), because temporal finite differences require contiguous tracked identities; a false-negative gap has no meaningful ``acceleration'' since the hand simply has no prediction for that frame, and interpolating or padding would conflate detection errors with smoothness.

\subsection{Prediction--Ground-Truth Alignment}
\label{sec:supp-alignment}

Each frame may contain zero, one, or two ground-truth hands (left, right). We adopt an IoU-based matching protocol that handles both our fixed-slot decoder and detector-based baselines uniformly.

For each frame, we compute 2D bounding boxes of both predicted and ground-truth MANO meshes (projected via pinhole intrinsics), dilating ground-truth boxes by 10\% to account for minor misalignment. Predictions are matched to ground-truth hands greedily by descending IoU, subject to a handedness constraint: a match is valid only if the predicted and ground-truth handedness labels agree and their bounding-box IoU exceeds $\tau_{\mathrm{IoU}}{=}0.1$. The threshold is deliberately permissive so that ambiguous near-miss predictions are still credited to the matching ground-truth hand rather than counted twice (as a false negative on the ground-truth side and a false positive on the prediction side). For our method, the left slot ($s=\mathrm{L}$) predicts the left hand and the right slot ($s=\mathrm{R}$) the right hand, so matching is deterministic. For baselines that produce an arbitrary number of predictions per frame, each prediction carries a handedness label from the detector or regressor; if multiple predictions match the same ground-truth hand, the one with the highest IoU is kept.

A matched prediction is a true positive (TP) if its hand-presence probability exceeds $0.5$. For our method we threshold the sigmoid of the on-screen visibility logit, $\sigma(\hat e) > 0.5$; for baselines that emit explicit per-prediction confidence (HaMeR, WiLoR, Hamba, OmniHands, InterWild), we use their reported probability; for tracker-based pipelines that emit an unconditional bounding box per tracked hand (HaWoR, Dyn-HaMR), we treat every emitted prediction as positive so the protocol does not penalize them for the absence of a presence head. A ground-truth hand with no valid match is a false negative (FN); a prediction with no valid match is a false positive (FP). We do not tune any baseline's operating point: each method runs at its own default confidence (or, for tracker-based pipelines, its emitted set), and the same $\tau_{\mathrm{IoU}}{=}0.1$ applies to every method, so the comparison reflects each method's intended deployment rather than a threshold we selected.

\paragraph{Off-screen filtering.}
A ground-truth hand is considered off-screen when none of its 21 MANO joints projects into the image plane $[0, W) \times [0, H)$ with $z > 0.01$\,m. The criterion uses the 21 skeletal joints rather than the 778 mesh vertices because MPJPE and EPE measure joint geometry directly, and a hand with only a sliver of fingertip mesh in frame would otherwise count the same as a fully visible hand. Under the off-screen-exclusion policy used throughout the paper, off-screen ground-truth hands are excluded from all metrics (not only 2D), and predictions matched to off-screen hands are excluded as well. This prevents penalizing methods for failing to detect hands that are entirely outside the field of view, while keeping the criterion ground-truth-only and method-independent.

\subsection{Coverage-Aware Evaluation Protocol}
\label{sec:supp-penalty}

Standard hand-pose benchmarks compute pose metrics (MPJPE, PA-MPJPE, etc.) only on true-positive detections, silently discarding false negatives. This rewards conservative detectors that skip ambiguous frames: a method detecting $85\%$ of hands with excellent per-prediction accuracy can appear to outperform a method detecting $97\%$ of hands with slightly lower per-prediction accuracy. To address this, we adopt a \emph{coverage-aware evaluation protocol} that includes every ground-truth hand instance---whether matched or missed---in every pose metric. We mark metrics computed under it with a ``-p'' subscript (for \emph{presence}: every present ground-truth hand is scored).

For each false negative, we substitute the \emph{canonical MANO placeholder}: the identity-rotation, zero-pose, mean-shape MANO output placed at the camera origin,
\begin{equation}
\widehat{\mathbf{h}}^{\mathrm{can}} = \bigl(\hat R = \mathbf{I}_3,\; \widehat{\boldsymbol\theta} = \mathbf{I}_3^{\otimes 15},\; \widehat{\boldsymbol\beta} = \mathbf{0},\; \widehat{\mathbf t} = \mathbf{0}\bigr)\,.
\label{eq:canonical}
\end{equation}
The canonical placeholder produces a per-sample MPJPE of about $132$\,mm on ARCTIC. Because MPJPE is root-relative (Eq.~\ref{eq:mpjpe}), absolute depth cancels out: the $132$\,mm reflects only the per-joint displacement between the canonical rest-pose articulation (fingers extended) and a typical articulated ground-truth hand once both have been wrist-anchored. The coverage-aware metric is then
\begin{equation}
\mathrm{metric}_{\text{-}\mathrm{p}} = \frac{\displaystyle\sum_{i \in \mathrm{TP}} e_i \;+\; \sum_{i \in \mathrm{FN}} e_i^{\mathrm{can}}}
{n_{\mathrm{TP}} + n_{\mathrm{FN}}}\,,
\label{eq:penalty}
\end{equation}
where $e_i$ is the per-sample error for matched predictions and $e_i^{\mathrm{can}}$ is the error of the canonical placeholder against the $i$-th missed ground-truth hand. The placeholder error is computed per sample (not a fixed constant), because it depends on the ground-truth hand's actual pose, orientation, and position. Per-metric placeholder definitions are given alongside each metric in \S\ref{sec:supp-metrics}.

This protocol produces a single number per metric that jointly reflects detection coverage and pose accuracy: a method that misses hands incurs a large, deterministic placeholder cost, while a method that detects all hands but predicts poorly also scores badly. Because the placeholder error is substantial relative to typical true-positive error, recall failures are reflected in the metric rather than hidden by it. This is strongest for the metrics whose placeholder greatly exceeds a typical true-positive error (MPJPE-p, PA-MPJPE-p, EPE-p, GO-p). Camera translation is the mildest case, but not a free pass for missed hands: the placeholder is the camera-to-hand distance $\|\mathbf{t}^\star\|$ ($\approx\!0.5$--$1.5$\,m across our benchmarks), roughly an order of magnitude larger than the detected-hand translation error of a well-localized method ($\approx\!0.04$--$0.12$\,m), so a missed hand is penalized substantially here as well. The placeholder stops adding cost beyond a bad detection only when a method's \emph{detected}-hand translation is itself catastrophically wrong---its error already approaching $\|\mathbf{t}^\star\|$, as for a depth-collapsing baseline---and such a method is already heavily penalized on the hands it does detect. We therefore treat CT-p as a valid coverage-aware translation metric, while still corroborating the detection story with the recall and F1 metrics.

\section{Per-Side Hand Detection Analysis}
\label{sec:supp-detection}

Tab.~\ref{tab:detection-summary} reports per-side (left, right, total) hand detection F1 for all baselines across the three benchmarks, isolating left--right confusion as asymmetric per-side F1 rather than as a separate handedness metric.

\begin{table}[!htbp]
\caption{\textbf{Per-side hand detection F1.} Left-hand (L), right-hand (R), and combined (Tot) detection F1 for the eight baselines and ViDiHand on the three benchmarks. Higher is better.}
\label{tab:detection-summary}
\centering
\resizebox{\textwidth}{!}{%
\begin{tabular}{l ccc ccc ccc}
\toprule
& \multicolumn{3}{c}{ARCTIC} & \multicolumn{3}{c}{HOT3D} & \multicolumn{3}{c}{HOI4D} \\
\cmidrule(lr){2-4} \cmidrule(lr){5-7} \cmidrule(lr){8-10}
Method & L & R & Tot & L & R & Tot & L & R & Tot \\
\midrule
InterWild~\cite{interwild_cvpr23}  & 0.955 & 0.964 & 0.959 & 0.802 & 0.925 & 0.868 & 0.473 & 0.954 & 0.864 \\
HaMeR~\cite{hamer_cvpr24}         & 0.949 & 0.964 & 0.957 & 0.826 & 0.935 & 0.883 & 0.473 & 0.956 & 0.864 \\
Hamba~\cite{hamba_nips24}          & 0.937 & 0.946 & 0.941 & 0.796 & 0.903 & 0.853 & 0.473 & 0.935 & 0.849 \\
WildHands~\cite{wildhands_eccv24}  & 0.953 & 0.966 & 0.960 & 0.797 & 0.885 & 0.844 & 0.470 & 0.956 & 0.864 \\
OmniHands~\cite{omnihands_arxiv24} & 0.945 & 0.961 & 0.954 & 0.810 & 0.921 & 0.868 & 0.385 & 0.962 & 0.834 \\
WiLoR~\cite{wilor_cvpr25}         & 0.975 & 0.974 & 0.974 & 0.921 & 0.949 & 0.937 & 0.893 & 0.982 & 0.972 \\
Dyn-HaMR~\cite{dynhamr_cvpr25}    & 0.951 & 0.951 & 0.951 & 0.735 & 0.856 & 0.802 & 0.790 & 0.851 & 0.845 \\
HaWoR~\cite{hawor_cvpr25}         & 0.906 & 0.883 & 0.895 & 0.620 & 0.682 & 0.654 & 0.837 & 0.928 & 0.919 \\
\rowcolor{oursrow} \textbf{ViDiHand (Ours)} & \textbf{1.000} & \textbf{0.999} & \textbf{0.999} & \textbf{0.981} & \textbf{0.985} & \textbf{0.983} & \textbf{0.986} & \textbf{0.991} & \textbf{0.990} \\
\bottomrule
\end{tabular}%
}
\end{table}

The asymmetry is most severe on HOI4D, where five baselines fall to a left-hand F1 between $0.385$ and $0.473$: most clips depict single-handed manipulation, and these pipelines emit a spurious prediction on the absent side. \nickname{} maintains per-side F1 at or above $0.981$ on every dataset.

\section{Pairwise Comparison on Co-Predicted Hands}
\label{sec:supp-tponly}

For completeness, Tabs.~\ref{tab:arctic-tponly}--\ref{tab:hoi4d-tponly} report a pairwise pose comparison restricted to co-predicted hands, in the conventional style of prior hand-pose papers that aggregates pose only over predicted hands and lets missed hands contribute nothing. The eight baselines use heterogeneous detection front-ends: per-frame ViT or CNN crops gated by an external hand detector (HaMeR, WiLoR, Hamba, WildHands), a multi-hand transformer with learned queries (OmniHands, InterWild), DROID-SLAM tracking (HaWoR), and 4D biomechanical optimization (Dyn-HaMR). Each method therefore has its own set of predicted hands, and intersecting these eight sets reduces the comparison to the lowest-recall baseline---for instance, HaWoR's $49.9\%$ recall on HOT3D would silently discard half the dataset and erase the recall advantage of every stronger detector. To keep each comparison head-to-head while accommodating heterogeneous detectors, we adopt the \emph{pairwise} variant of the protocol: for each baseline $B$, both ViDiHand and $B$ are scored on the same set of (frame, hand-side) pairs where both methods emit a same-side prediction and the ground-truth hand is on screen. Each cell is reported as $B/\mathrm{ours}$, both averaged on this shared sample set, and bold marks the better of the two numbers; because the sample set varies per row, the ViDiHand number varies slightly across rows. Because the aim here is to compare pose accuracy rather than detection, predictions are paired to ground truth by hand identity: every (frame, hand-side) at which both methods commit a same-side prediction contributes, with no spatial-overlap gate. This is deliberate---a spatial gate would discard predictions that fall on the correct hand but are imperfectly localized and would fold localization accuracy into the pose score, whereas here each method is judged on the pose of every hand it commits to. This set consequently differs from the true-positive set of the main paper's coverage-aware protocol, which matches predictions to ground truth by bounding-box overlap and charges a bounded placeholder for every unmatched hand: a method's on-the-correct-hand-but-loosely-localized predictions are scored here yet fall below that overlap threshold and are charged the placeholder as false-negatives there instead. The two sets therefore differ by more than the recall gap, and their per-metric values are not related by a recall-weighted bound---most visibly for camera translation, which is dominated by spatial placement and can rank methods differently across the two tables. Identity-based matching, unlike the overlap gate, does not discard a prediction whose depth regression has collapsed to a physically impossible value. On a handful of HOI4D frames OmniHands places a hand thousands of meters from the camera---up to ${\sim}4000$\,m, its $21$ joints projecting to a single pixel---which inflates its mean co-predicted camera translation. The tables above report these values \emph{unclamped}. As a robustness check, bounding each method's per-hand camera translation at $100$\,m (in the spirit of the image-diagonal EPE2D clamp, \S\ref{sec:supp-metrics}) moves only OmniHands: its co-predicted CT falls from $0.070/0.119/0.183$\,m on ARCTIC/HOT3D/held-out HOI4D to $0.064/0.114/0.070$\,m, and the threshold is immaterial (a $30$\,m bound gives $0.059/0.106/0.068$\,m). Every other method is unchanged, as OmniHands is the only one with any hand beyond $30$\,m---$4$, $12$, and $1$ (frame, hand-side) pairs on the three splits, respectively. Under this bound OmniHands' held-out HOI4D CT ($0.070$\,m) drops below \nickname{}'s ($0.111$\,m), flipping that single cell; we keep the unclamped value in the table and disclose the bounded one here so both are visible. The main-paper coverage-aware protocol needs no such bound: there the same one-pixel degenerate prediction fails the IoU gate and its ground-truth hand instead receives the bounded placeholder $\|\mathbf{t}^\star\|$, so the outlier never enters the average.

\begin{table}[!htbp]
\caption{\textbf{Pairwise comparison on co-predicted hands, ARCTIC.} For each baseline (row) and metric (column), the cell reports baseline\,/\,ViDiHand, computed only on (frame, hand-side) pairs where both methods emit a same-side prediction and the ground-truth hand is on screen. Bold marks the better value in each cell.}
\label{tab:arctic-tponly}
\centering
\resizebox{\textwidth}{!}{%
\begin{tabular}{l ccccc}
\toprule
\multirow{2}{*}{\textbf{Method}} & \multicolumn{2}{c}{\textbf{3D Pose}} & \multicolumn{3}{c}{\textbf{Orient.\ \& Position}} \\
\cmidrule(lr){2-3} \cmidrule(lr){4-6}
 & MPJPE (mm) & PA-MPJPE (mm) & EPE2D (px) & GO ($^\circ$) & CT (m) \\
\midrule
InterWild~\cite{interwild_cvpr23}  & 27.46 / \textbf{20.92} & 11.05 / \textbf{9.63} & 17.23 / \textbf{11.16} & 19.87 / \textbf{14.03} & 0.072 / \textbf{0.046} \\
HaMeR~\cite{hamer_cvpr24}          & 26.16 / \textbf{20.93} & \textbf{9.60} / 9.64 & 29.49 / \textbf{11.17} & 19.84 / \textbf{14.04} & 0.070 / \textbf{0.046} \\
Hamba~\cite{hamba_nips24}          & 25.74 / \textbf{20.65} & \textbf{9.41} / 9.58 & 29.44 / \textbf{10.89} & 19.03 / \textbf{13.92} & 0.070 / \textbf{0.046} \\
WildHands~\cite{wildhands_eccv24}  & 21.83 / \textbf{20.95} & \textbf{9.21} / 9.64   & 15.85 / \textbf{11.19} & 16.57 / \textbf{14.05} & \textbf{0.032} / 0.046 \\
OmniHands~\cite{omnihands_arxiv24} & 26.93 / \textbf{21.04} & 9.73 / \textbf{9.65} & 19.55 / \textbf{11.32} & 19.94 / \textbf{14.11} & 0.070 / \textbf{0.046} \\
WiLoR~\cite{wilor_cvpr25}          & \textbf{18.05} / 20.87 & \textbf{7.35} / 9.64 & 34.86 / \textbf{10.98} & \textbf{11.90} / 14.06  & 0.050 / \textbf{0.046} \\
Dyn-HaMR~\cite{dynhamr_cvpr25}     & 22.10 / \textbf{21.10} & \textbf{9.61} / 9.67 & 30.82 / \textbf{11.32} & 17.13 / \textbf{14.23} & 0.083 / \textbf{0.045} \\
HaWoR~\cite{hawor_cvpr25}          & 32.75 / \textbf{20.31} & \textbf{9.12} / 9.51 & 30.46 / \textbf{10.54} & 23.26 / \textbf{13.82} & 0.058 / \textbf{0.045} \\
\bottomrule
\end{tabular}%
}
\end{table}

\begin{table}[!htbp]
\caption{\textbf{Pairwise comparison on co-predicted hands, HOT3D.} Same protocol as Tab.~\ref{tab:arctic-tponly}: each cell reports baseline\,/\,ViDiHand on (frame, hand-side) pairs where both methods emit a same-side prediction and the ground-truth hand is on screen, with bold marking the better value.}
\label{tab:hot3d-tponly}
\centering
\resizebox{\textwidth}{!}{%
\begin{tabular}{l ccccc}
\toprule
\multirow{2}{*}{\textbf{Method}} & \multicolumn{2}{c}{\textbf{3D Pose}} & \multicolumn{3}{c}{\textbf{Orient.\ \& Position}} \\
\cmidrule(lr){2-3} \cmidrule(lr){4-6}
 & MPJPE (mm) & PA-MPJPE (mm) & EPE2D (px) & GO ($^\circ$) & CT (m) \\
\midrule
InterWild~\cite{interwild_cvpr23}  & 73.57 / \textbf{17.71} & 11.16 / \textbf{7.83} & 21.33 / \textbf{6.61} & 50.48 / \textbf{11.84} & 0.226 / \textbf{0.027} \\
HaMeR~\cite{hamer_cvpr24}          & 63.88 / \textbf{17.69} & 10.32 / \textbf{7.82} & 21.23 / \textbf{6.61} & 42.34 / \textbf{11.83} & 0.090 / \textbf{0.027} \\
Hamba~\cite{hamba_nips24}          & 61.83 / \textbf{17.17} & 9.51 / \textbf{7.71} & 20.92 / \textbf{6.37} & 40.99 / \textbf{11.40} & 0.076 / \textbf{0.026} \\
WildHands~\cite{wildhands_eccv24}  & 41.06 / \textbf{17.68} & 12.46 / \textbf{7.82} & 51.41 / \textbf{6.61} & 39.01 / \textbf{11.83} & 0.116 / \textbf{0.027} \\
OmniHands~\cite{omnihands_arxiv24} & 56.42 / \textbf{17.74} & 10.22 / \textbf{7.84} & 24.72 / \textbf{6.63} & 38.99 / \textbf{11.89} & 0.119 / \textbf{0.027} \\
WiLoR~\cite{wilor_cvpr25}          & 19.23 / \textbf{17.37} & \textbf{6.82} / 7.77 & 27.14 / \textbf{6.62} & 12.24 / \textbf{11.57} & 0.054 / \textbf{0.026} \\
Dyn-HaMR~\cite{dynhamr_cvpr25}     & 64.94 / \textbf{17.98} & 17.88 / \textbf{7.82} & 91.27 / \textbf{6.78} & 24.44 / \textbf{12.14} & 0.587 / \textbf{0.027} \\
HaWoR~\cite{hawor_cvpr25}          & 19.63 / \textbf{16.48} & \textbf{7.26} / 7.47 & 12.27 / \textbf{6.19} & 13.27 / \textbf{10.91} & 0.051 / \textbf{0.027} \\
\bottomrule
\end{tabular}%
}
\end{table}

\begin{table}[!htbp]
\caption{\textbf{Pairwise comparison on co-predicted hands, held-out HOI4D.} Same protocol as Tab.~\ref{tab:arctic-tponly}, evaluated with the main-paper ViDiHand model, for which HOI4D is held out (trained only on ARCTIC and HOT3D, with no explicit HOI4D supervision). Each cell reports baseline\,/\,ViDiHand, with bold marking the better value.}
\label{tab:hoi4d-tponly}
\centering
\resizebox{\textwidth}{!}{%
\begin{tabular}{l ccccc}
\toprule
\multirow{2}{*}{\textbf{Method}} & \multicolumn{2}{c}{\textbf{3D Pose}} & \multicolumn{3}{c}{\textbf{Orient.\ \& Position}} \\
\cmidrule(lr){2-3} \cmidrule(lr){4-6}
 & MPJPE (mm) & PA-MPJPE (mm) & EPE2D (px) & GO ($^\circ$) & CT (m) \\
\midrule
InterWild~\cite{interwild_cvpr23}  & 46.09 / \textbf{28.82} & 13.18 / \textbf{12.86} & \textbf{15.02} / 16.79 & 32.90 / \textbf{21.87} & 0.187 / \textbf{0.111} \\
HaMeR~\cite{hamer_cvpr24}          & 36.77 / \textbf{28.82} & \textbf{11.95} / 12.86 & \textbf{15.35} / 16.79 & 24.08 / \textbf{21.87} & 0.142 / \textbf{0.111} \\
Hamba~\cite{hamba_nips24}          & 35.33 / \textbf{28.51} & \textbf{11.32} / 12.74 & \textbf{15.58} / 16.71 & 23.25 / \textbf{21.53} & 0.139 / \textbf{0.111} \\
WildHands~\cite{wildhands_eccv24}  & 38.11 / \textbf{28.82} & 14.24 / \textbf{12.87} & 19.06 / \textbf{16.80} & 37.28 / \textbf{21.87} & 0.113 / \textbf{0.111} \\
OmniHands~\cite{omnihands_arxiv24} & 38.04 / \textbf{28.80} & \textbf{10.71} / 12.85 & 18.63 / \textbf{16.82} & 26.95 / \textbf{21.82} & 0.183 / \textbf{0.111} \\
WiLoR~\cite{wilor_cvpr25}          & 30.04 / \textbf{29.06} & \textbf{10.55} / 12.88 & 17.11 / \textbf{17.08} & \textbf{21.09} / 22.19 & \textbf{0.093} / 0.111 \\
Dyn-HaMR~\cite{dynhamr_cvpr25}     & 31.72 / \textbf{29.30} & \textbf{12.57} / 12.88 & 19.98 / \textbf{17.25} & 23.29 / \textbf{22.51} & 0.190 / \textbf{0.112} \\
HaWoR~\cite{hawor_cvpr25}          & 32.99 / \textbf{28.63} & \textbf{11.48} / 12.68 & \textbf{12.64} / 16.85 & 26.15 / \textbf{22.03} & \textbf{0.052} / 0.112 \\
\bottomrule
\end{tabular}%
}
\end{table}

\paragraph{Pairwise win rate.}
This co-predicted-hands protocol is structurally favorable to baselines, because by construction it excludes precisely the difficult hands (severe hand--object or hand--hand occlusion, field-of-view truncation, motion blur, fisheye periphery) where some baseline misses the detection---exactly the regimes the video world model's internal representation is designed to handle. Despite this structural advantage to baselines, ViDiHand wins $96$ of $120$ $B/\mathrm{ours}$ cells: $31$ of $40$ on ARCTIC, $38$ of $40$ on HOT3D, and $27$ of $40$ on the \emph{held-out} HOI4D split. On the in-distribution benchmarks the exceptions concentrate on Procrustes-aligned MPJPE, which factors out global rotation, translation, and scale and thus scores only local articulation; on the subset of hands they successfully detect, the strongest baselines---both crop-based regressors and the SLAM-refined world-frame methods Dyn-HaMR and HaWoR---match or surpass ViDiHand under this alignment. On the held-out HOI4D split the baseline wins are broader, spreading across Procrustes-aligned MPJPE, 2D reprojection, and camera translation, consistent with the wider cross-dataset gap noted below: on held-out HOI4D, evaluated with no explicit HOI4D supervision, as is every baseline, ViDiHand still leads all eight methods on absolute MPJPE and, against nearly all of them, on global orientation, and it attains the highest detection coverage of any method (main paper); the specialized single-frame regressors are competitive on the alignment-invariant pose and 2D-reprojection metrics, and---as the main paper reports for CT-p---several baselines reach a lower absolute camera-translation error on this unfamiliar split, leaving ViDiHand mid-ranked on that one metric. Finally, the co-predicted-hands comparison here and the coverage-aware metrics of the main paper are distinct protocols---the former scores only the hands two methods jointly detect, the latter every ground-truth hand with a deterministic placeholder for the missed ones (\S\ref{sec:supp-penalty})---so their rankings need not coincide, and neither bounds the other. This is clearest for camera translation. Two effects make the two numbers differ, and either can make the co-predicted value the larger one. First, they score different hand sets---only jointly-detected hands here, every ground-truth hand there. Second, the coverage-aware value replaces each missed hand with the bounded placeholder $\|\mathbf{t}^\star\|$ (the camera-to-hand distance), which for a method whose detected-hand translation error is very large is \emph{cheaper} than that error and pulls the coverage-aware mean down. On HOT3D both cases appear: Dyn-HaMR ($0.587$ vs.\ $0.571$\,m) is the second---its detected-hand CT exceeds the placeholder---while InterWild ($0.226$ vs.\ $0.213$\,m) is the first, since there the placeholder (the head-mounted camera-to-hand distance) is \emph{larger} than its error, so the small gap comes from the differing hand sets rather than the placeholder. Either way, a co-predicted CT can exceed a method's coverage-aware CT-p with no degenerate prediction involved. These are not degenerate-depth outliers; they are the expected consequence of the two protocols measuring different things, and the $100$\,m bound (which touches only OmniHands) neither creates nor removes them. The coverage-aware protocol of \S\ref{sec:supp-penalty}, which folds every missed hand back in, is the one that reflects end-to-end usability.

\paragraph{Capacity reference: HOI4D in-domain.}
Every comparison above uses the held-out ViDiHand of the main paper, which receives no explicit HOI4D supervision at any training stage. As the main-paper limitations note, this refers to explicit supervision only: \nickname{} builds on an internet-scale pretrained video backbone whose corpus cannot be audited for incidental exposure, so we describe HOI4D as held out from explicit supervision rather than provably unseen. For completeness, Tab.~\ref{tab:hoi4d-train-vs-heldout} additionally reports the metrics of the identical readout trained with HOI4D in-domain, and Tab.~\ref{tab:hoi4d-tponly-indomain} its pairwise comparison, to indicate the accuracy the architecture reaches once the distribution is in-domain---the same regime it attains on ARCTIC and HOT3D. These in-domain numbers are \emph{not} a held-out comparison (the baselines are never trained on HOI4D) and are not used in any claim against baselines; the held-out model of Tab.~\ref{tab:hoi4d-tponly} is the one on which all baseline comparisons rest.

\begin{table}[!htbp]
\caption{\textbf{HOI4D: held-out versus in-domain ViDiHand.} Main-paper metrics (coverage-aware protocol, \S\ref{sec:supp-penalty}) on HOI4D for the held-out ViDiHand (the main-paper model; no explicit HOI4D supervision during training) and for the identical readout trained with HOI4D in-domain. Detection metrics are higher-better; the rest are lower-better. The in-domain readout (\textbf{bold}) reaches the same accuracy regime as on the in-distribution benchmarks; it is reported only to indicate the architecture's capacity, not as a baseline comparison.}
\label{tab:hoi4d-train-vs-heldout}
\centering
\resizebox{\textwidth}{!}{%
\begin{tabular}{l ccc cc ccc c}
\toprule
\multirow{2}{*}{\textbf{ViDiHand on HOI4D}} & \multicolumn{3}{c}{\textbf{Detection}} & \multicolumn{2}{c}{\textbf{3D Pose}} & \multicolumn{3}{c}{\textbf{Orient.\ \& Position}} & \textbf{Temporal} \\
\cmidrule(lr){2-4}\cmidrule(lr){5-6}\cmidrule(lr){7-9}\cmidrule(lr){10-10}
 & FAcc & Recall & F1 & MPJPE-p & PA-MPJPE-p & EPE-p & GO-p & CT-p & Jitter \\
\midrule
Held-out (not HOI4D-trained)       & 0.984 & 0.991 & 0.990 & 30.090 & 13.960 & 24.460 & 23.420 & 0.117 & 4.010 \\
In-domain (HOI4D in training) & \textbf{0.993} & \textbf{0.996} & \textbf{0.996} & \textbf{15.674} & \textbf{7.253} & \textbf{9.693} & \textbf{13.015} & \textbf{0.025} & \textbf{2.642} \\
\bottomrule
\end{tabular}%
}
\end{table}

\begin{table}[!htbp]
\caption{\textbf{Pairwise comparison on co-predicted hands, in-domain HOI4D.} As Tab.~\ref{tab:hoi4d-tponly}, but with ViDiHand trained with HOI4D in-domain while the baselines remain untrained on HOI4D; this is a capacity reference, \emph{not} a held-out comparison. Each cell reports baseline\,/\,ViDiHand, with bold marking the better value.}
\label{tab:hoi4d-tponly-indomain}
\centering
\resizebox{\textwidth}{!}{%
\begin{tabular}{l ccccc}
\toprule
\multirow{2}{*}{\textbf{Method}} & \multicolumn{2}{c}{\textbf{3D Pose}} & \multicolumn{3}{c}{\textbf{Orient.\ \& Position}} \\
\cmidrule(lr){2-3} \cmidrule(lr){4-6}
 & MPJPE (mm) & PA-MPJPE (mm) & EPE2D (px) & GO ($^\circ$) & CT (m) \\
\midrule
InterWild~\cite{interwild_cvpr23}  & 46.06 / \textbf{15.05} & 13.17 / \textbf{6.68} & 15.01 / \textbf{7.58} & 32.88 / \textbf{12.26} & 0.187 / \textbf{0.022} \\
HaMeR~\cite{hamer_cvpr24}          & 36.76 / \textbf{15.05} & 11.94 / \textbf{6.68} & 15.35 / \textbf{7.58} & 24.07 / \textbf{12.26} & 0.142 / \textbf{0.022} \\
Hamba~\cite{hamba_nips24}          & 35.31 / \textbf{14.92} & 11.32 / \textbf{6.60} & 15.58 / \textbf{7.52} & 23.24 / \textbf{12.09} & 0.139 / \textbf{0.022} \\
WildHands~\cite{wildhands_eccv24}  & 38.09 / \textbf{15.05} & 14.23 / \textbf{6.68} & 19.05 / \textbf{7.58} & 37.30 / \textbf{12.26} & 0.113 / \textbf{0.022} \\
OmniHands~\cite{omnihands_arxiv24} & 38.01 / \textbf{15.04} & 10.71 / \textbf{6.68} & 18.61 / \textbf{7.59} & 26.94 / \textbf{12.25} & 0.183 / \textbf{0.022} \\
WiLoR~\cite{wilor_cvpr25}          & 30.03 / \textbf{15.11} & 10.55 / \textbf{6.74} & 17.11 / \textbf{7.64} & 21.08 / \textbf{12.43} & 0.093 / \textbf{0.022} \\
Dyn-HaMR~\cite{dynhamr_cvpr25}     & 31.62 / \textbf{15.25} & 12.56 / \textbf{6.78} & 19.96 / \textbf{7.69} & 23.22 / \textbf{12.65} & 0.190 / \textbf{0.022} \\
HaWoR~\cite{hawor_cvpr25}          & 32.96 / \textbf{14.95} & 11.48 / \textbf{6.69} & 12.64 / \textbf{7.57} & 26.14 / \textbf{12.22} & 0.052 / \textbf{0.022} \\
\bottomrule
\end{tabular}%
}
\end{table}

\section{Additional Qualitative Comparisons}
\label{sec:supp-qualitative}

\begin{figure}[htbp]
  \centering
  \includegraphics[width=\textwidth]{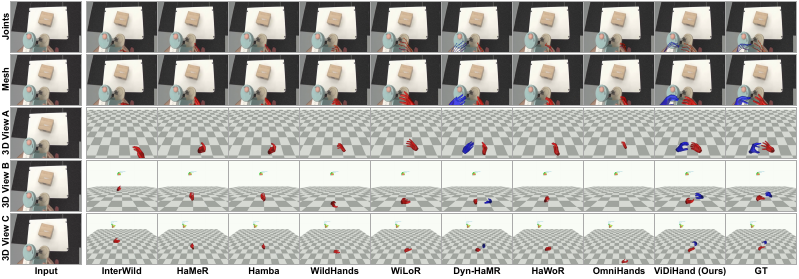}
  \caption{\textbf{Qualitative comparison on ARCTIC.} Bimanual mixer manipulation where both hands are almost entirely hidden behind the mixer body they grasp and cropped by the frame's bottom edge, leaving only fingertips visible. InterWild, HaMeR, Hamba, WildHands, WiLoR, HaWoR, and OmniHands each recover only one hand and miss the other entirely; Dyn-HaMR alone outputs both hands, but with distorted articulation and mismatched relative positioning. \nickname{} alone recovers the full bimanual configuration, correctly separating both hands' identity and depth.}
  \label{fig:supp-arctic-mixer}
\end{figure}

\begin{figure}[htbp]
  \centering
  \includegraphics[width=\textwidth]{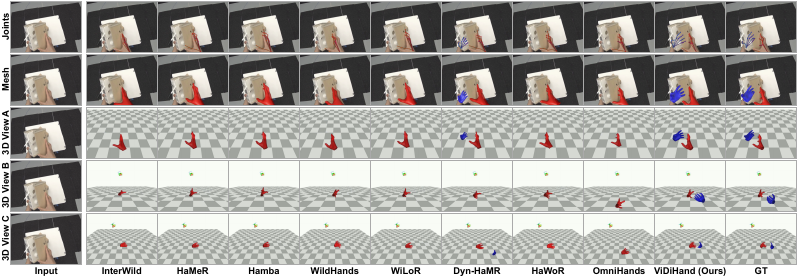}
  \caption{\textbf{Qualitative comparison on ARCTIC.} Bimanual manipulation of an articulated espresso machine, where the right hand operates it while the left hand is occluded behind the machine. InterWild, HaMeR, Hamba, WildHands, WiLoR, HaWoR, and OmniHands recover only the visible right hand, never the occluded left one; Dyn-HaMR alone predicts a left hand, but places it at the wrong depth, where it vanishes entirely in View B. \nickname{} alone keeps the occluded hand's position and articulation stable across all three views.}
  \label{fig:supp-arctic-espresso}
\end{figure}

\begin{figure}[htbp]
  \centering
  \includegraphics[width=\textwidth]{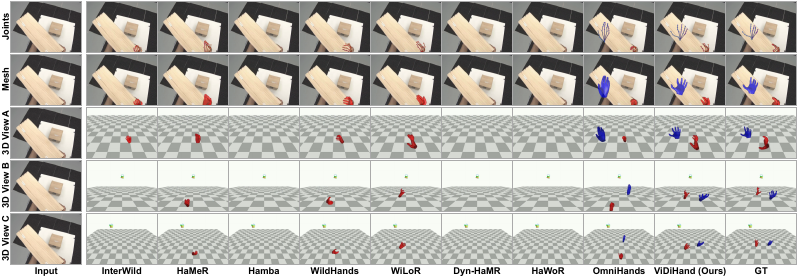}
  \caption{\textbf{Qualitative comparison on ARCTIC.} Both hands grasp the box from underneath: the left hand is entirely hidden beneath the box, while the right hand is likewise hidden beneath the box's far end and further cropped by the bottom frame edge, leaving only a fingertip visible. InterWild, HaMeR, WildHands, and WiLoR recover only the partially-visible right hand and miss the fully occluded left one; Hamba, Dyn-HaMR, and HaWoR miss both hands entirely; OmniHands detects both hands but with inconsistent scale and an implausible finger pose on the occluded left hand, while its right-hand mesh collapses to a blob in the novel views. \nickname{} reconstructs both hands with correct relative placement despite one hand being entirely invisible and the other occluded and clipped by the frame.}
  \label{fig:supp-main-arctic-box}
\end{figure}

\begin{figure}[htbp]
  \centering
  \includegraphics[width=\textwidth]{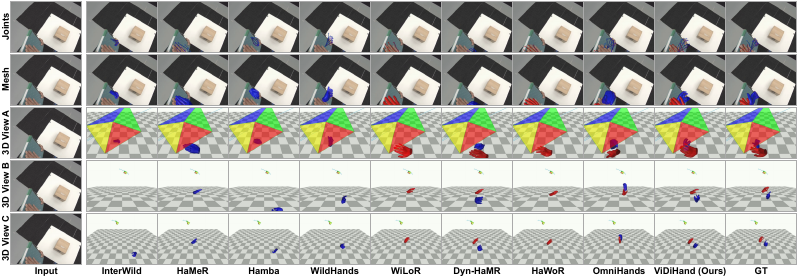}
  \caption{\textbf{Qualitative comparison on ARCTIC.} A grasp where one hand rests with fingers splayed flat and fully visible on the notebook's cover, while the other grips its near corner from behind, almost entirely hidden with only a sliver of fingers peeking over the edge. HaMeR drops the hidden corner-gripping hand and renders the visible one as a fused, fingerless mass; Hamba and InterWild instead drop the fully-visible hand and render the hidden one as a fingerless blob. OmniHands keeps both hands but tangles the corner-gripping hand into a self-crossing joint layout; \nickname{} recovers both hands consistently with the ground truth, qualitatively supporting its low 2D end-point error.}
  \label{fig:supp-main-arctic-notebook}
\end{figure}

\begin{figure}[htbp]
  \centering
  \includegraphics[width=\textwidth]{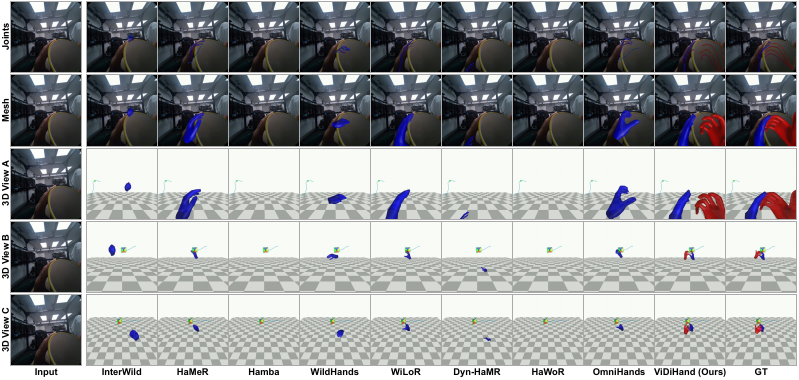}
  \caption{\textbf{Qualitative comparison on HOT3D.} Both hands grip a cup under the wide-angle fisheye in a dim scene: the left hand grips the near rim in full view, while the right hand wraps around the far side and is entirely hidden behind the cup itself. Hamba and HaWoR miss both hands entirely, InterWild's left-hand mesh collapses into a small disconnected blob floating off the rim, while HaMeR, WildHands, WiLoR, and OmniHands each recover only the visible left hand; \nickname{} alone reconstructs both hands, correctly inferring the fully occluded right hand to match GT.}
  \label{fig:supp-hot3d-83b755aa}
\end{figure}

\begin{figure}[htbp]
  \centering
  \includegraphics[width=\textwidth]{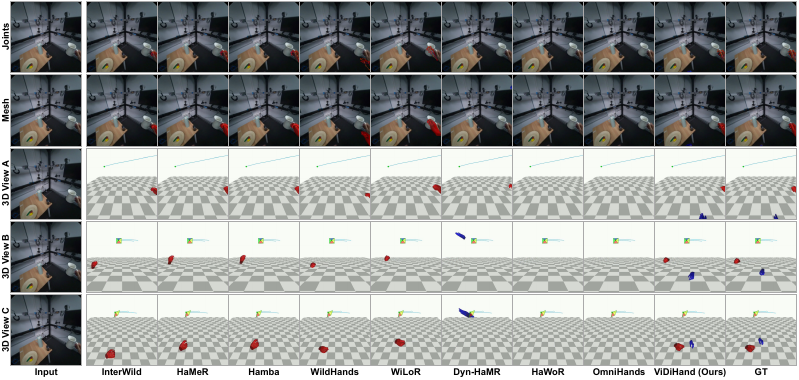}
  \caption{\textbf{Qualitative comparison on HOT3D.} A hand grips a pitcher near the periphery of the wide-angle view, mostly hidden behind it and heavily foreshortened there, which breaks crop-based detection. HaWoR and OmniHands detect no hand at all, and Dyn-HaMR's predicted hand collapses toward the camera instead of the correct depth. \nickname{}'s ray-space encoding keeps wrist orientation and position aligned with the ground truth across all three novel views, and alone recovers the near-invisible left hand that every baseline misses.}
  \label{fig:supp-hot3d-95eabc0a}
\end{figure}

\begin{figure}[htbp]
  \centering
  \includegraphics[width=\textwidth]{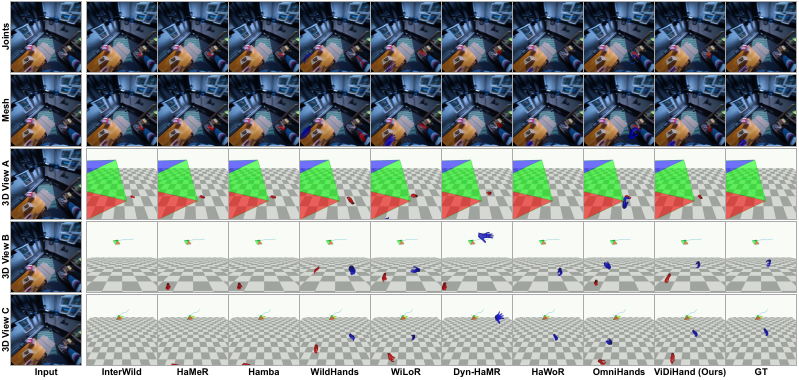}
  \caption{\textbf{Qualitative comparison on HOT3D.} Two hands are present in a difficult peripheral configuration: the tracked hand is motion-blurred and mostly cropped by the bottom frame edge, while a second hand reaches toward a shelf near the frame boundary. InterWild, HaMeR, and Hamba recover only the shelf hand and miss the cropped one, while HaWoR recovers only the cropped hand and misses the shelf hand entirely. WildHands and WiLoR predict both hands but with inaccurate position and articulation, OmniHands mislocalizes both hand predictions onto the shelf hand's position, and Dyn-HaMR's second hand floats implausibly close to the camera. \nickname{} alone recovers both hands with plausible position and articulation.}
  \label{fig:supp-hot3d-5766eae8}
\end{figure}

\begin{figure}[htbp]
  \centering
  \includegraphics[width=\textwidth]{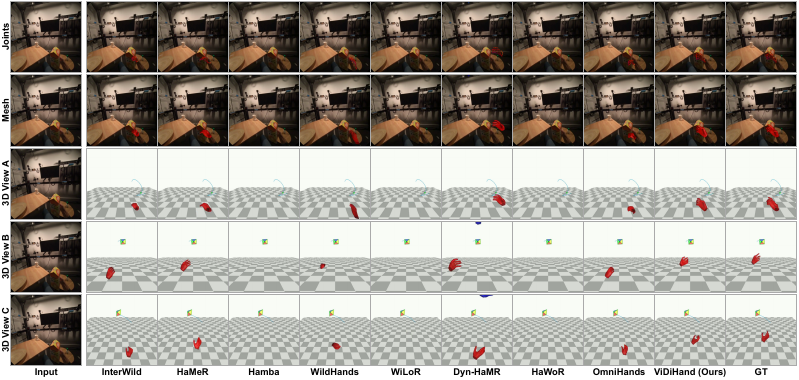}
  \caption{\textbf{Qualitative comparison on HOT3D.} A single-hand interaction under dim, high-dynamic-range lighting. Hamba, WiLoR, and HaWoR fail to detect the hand; InterWild, HaMeR, WildHands, and OmniHands detect it but with inaccurate articulation; and Dyn-HaMR's hand mesh is positioned wrong throughout, appearing far larger than ground truth in the side views. \nickname{} alone matches the ground-truth position, orientation, and articulation throughout.}
  \label{fig:supp-hot3d-0c2d00ed}
\end{figure}

\begin{figure}[htbp]
  \centering
  \includegraphics[width=\textwidth]{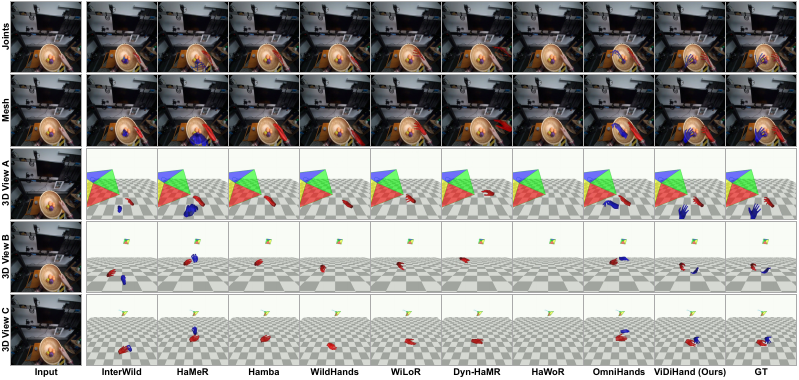}
  \caption{\textbf{Qualitative comparison on HOT3D.} Both hands grip a bowl: the left hand is severely occluded and invisible in the raw frame, while the right hand is only partially visible, its fingers curling over the rim. Hamba, WildHands, and WiLoR recover only the right hand and drop the left entirely; HaWoR drops both; and Dyn-HaMR's single detected hand drifts onto the background shelf. HaMeR and InterWild attempt the occluded left hand but collapse it into a shapeless, fingerless blob, while OmniHands recovers it with reversed orientation. Even where a right hand is recovered, most baselines still show inaccurate pose and position; \nickname{} alone keeps both slots correctly separated, articulated, and aligned with the ground truth.}
  \label{fig:supp-hot3d-76ea6d47}
\end{figure}

\begin{figure}[htbp]
  \centering
  \includegraphics[width=\textwidth]{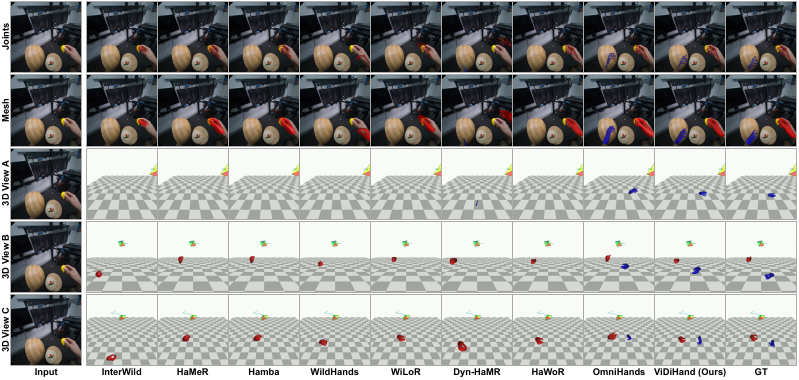}
  \caption{\textbf{Qualitative comparison on HOT3D.} Full nine-method expansion of the HOT3D example summarized in the main paper. At the wide-angle periphery, the left hand is almost entirely occluded behind the large bowl; every baseline except OmniHands fails to detect it, and OmniHands recovers it at the wrong position and depth with under-articulated fingers. \nickname{} matches the ground-truth position and articulation of both hands across all three novel views.}
  \label{fig:supp-main-hot3d-seg011}
\end{figure}

\begin{figure}[htbp]
  \centering
  \includegraphics[width=\textwidth]{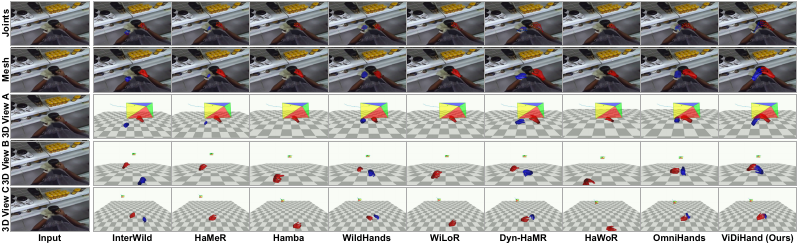}
  \caption{\textbf{In-the-wild qualitative comparison.} A bimanual grasp of a shelved canister, with the left hand occluded by cloth. Hamba, WiLoR, and HaWoR fail to detect the occluded left hand. InterWild and HaMeR collapse it into a small blob far below the grip, with a large depth offset. WildHands, Dyn-HaMR, and OmniHands place it near the correct grip point but still as a fingerless blob. \nickname{} alone recovers both hands with visually plausible position, depth, and articulation, coherently grasping the canister.}
  \label{fig:supp-wild-ropedia-seg001}
\end{figure}

\begin{figure}[htbp]
  \centering
  \includegraphics[width=\textwidth]{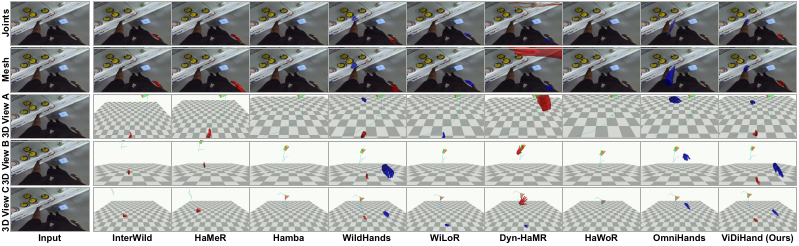}
  \caption{\textbf{In-the-wild qualitative comparison.} The right hand is clearly visible with fingers spread, while the left is almost entirely occluded at the shelf edge, with only the wrist showing. InterWild, HaMeR, Hamba, WiLoR, and HaWoR miss the occluded left hand; WildHands' blob lies flat along the shelf instead of curling into the gap, and OmniHands gets both its depth and position wrong; Dyn-HaMR hallucinates an oversized phantom hand at the frame's top edge. The visible right hand is often wrong too: InterWild and HaMeR curl its fingers into a tangled shape, WildHands' and Dyn-HaMR's meshes sit displaced below the fingers, and Hamba, HaWoR, and OmniHands miss it. \nickname{} alone recovers both hands with visually plausible position, depth, and articulation.}
  \label{fig:supp-wild-ropedia-seg005}
\end{figure}

\begin{figure}[htbp]
  \centering
  \includegraphics[width=\textwidth]{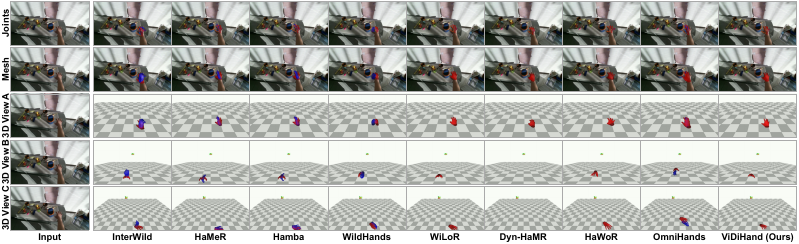}
  \caption{\textbf{In-the-wild qualitative comparison.} A single-hand grasp under dappled, uneven sunlight. InterWild, HaMeR, Hamba, and WildHands each mistake the lighting for a second hand, detecting the visible hand into both slots as duplicates; HaWoR and OmniHands recover a single hand at a visibly wrong rotation. WiLoR's pose is inaccurate; Dyn-HaMR's hand has the wrong depth and its fingers are bent noticeably more than the true pose, vanishing from two of the three novel views despite a plausible 2D projection. \nickname{} recovers the hand's position, depth, and articulation, and reports the second slot as absent.}
  \label{fig:supp-wild-hoi4d}
\end{figure}

Figs.~\ref{fig:supp-arctic-mixer}--\ref{fig:supp-wild-hoi4d} present per-frame qualitative comparisons spanning ARCTIC, HOT3D, and in-the-wild clips. Each figure shows a single frame across all eight baselines and ViDiHand, with five visualization rows: projected 2D joints, reprojected MANO mesh overlay, and three novel-viewpoint 3D renderings (Views A--C). For in-the-wild clips without ground-truth MANO, the GT column is omitted.

\clearpage
\section{Implementation Details}
\label{sec:supp-implementation}

This section consolidates the data, decoder architecture, losses, and two-stage training procedure that together specify the full pipeline. The video backbone is Wan2.1-VACE~\cite{wan,vace}, a $1.3$-billion-parameter video diffusion transformer (DiT) with a VACE branch that injects the egocentric input video as a conditioning signal. Throughout, $L_{15}$ refers to the 15th of the 30 transformer blocks; the normalized denoising step $\tau_k$ ($\tau{=}0$ pure noise, $\tau{=}1$ fully denoised) corresponds to $\tau_k$ in the main paper.

\subsection{Datasets and Cameras}
\label{sec:supp-datasets}

We briefly describe each dataset and then provide camera intrinsics and training composition (Tab.~\ref{tab:dataset-spec-supp}). All clips are sliced into consecutive $81$-frame segments; the Wan2.1 variational autoencoder temporally compresses each segment to $F_{\mathrm{lat}}{=}21$ latent frames, and the spatial patch grid after the height-to-$480$ preprocessing resize is $30{\times}42$ for ARCTIC, $30{\times}30$ for HOT3D, and $30{\times}54$ for HOI4D (the $854$-px HOI4D width is padded to $864$, the next multiple of the $16$-px patch size, before patchification).

\paragraph{EgoDex~\cite{egodex}.} A large-scale egocentric dataset for dexterous manipulation, containing diverse hand--object interactions across hundreds of scenes. It provides 3D joint annotations obtained via multi-view triangulation but does not include fitted MANO mesh parameters. We use EgoDex exclusively in Stage~1a for joint-overlay pretraining.

\paragraph{ARCTIC~\cite{arctic_cvpr23}.} A bimanual hand--object manipulation dataset captured from a chest-mounted camera with about $60^\circ$ horizontal field of view. Subjects interact with articulated objects (e.g., scissors, laptops, microwaves) using both hands, producing frequent and severe occlusion of multiple kinds: fingers wrap around object surfaces and disappear behind them (hand--object occlusion); the two hands grasp the same object from opposite sides so that one hand passes behind or fully occludes the other (hand--hand occlusion); and grasps routinely leave only a few fingertips visible. MANO parameters are obtained via multi-view motion capture.

\paragraph{HOT3D~\cite{hot3d_cvpr25}.} An egocentric dataset captured with Meta Aria glasses, whose wide-angle fisheye camera spans about $99^\circ$ horizontal field of view. We undistort each frame to the dataset's linear (pinhole) camera model ($f_x \approx 208$ after resize) and compute all projection---EPE-p, the ray-space encoding, and the mixed-projection solve---in that rectified space, so no residual radial distortion enters our metrics. Rectification removes the distortion but not the field of view: the view remains a $99^\circ$ wide-angle one, so hands at the periphery---where they frequently appear under egocentric reaching motions---stay strongly foreshortened and perspective-stretched, which is part of what makes detection on HOT3D harder than on ARCTIC. Indoor scenes routinely combine bright window or lamp regions with deep shadows in the same frame, producing high-dynamic-range lighting that shifts hand appearance across the frame. The head-mounted setup couples large head motion with rapid hand reaches, producing motion blur on both the camera and the hands. All eight baselines see lower detection F1 on HOT3D than on ARCTIC (Tab.~\ref{tab:detection-summary}). Since HOT3D does not release ground-truth MANO for its official test sequences, we randomly hold out $5\%$ of the validation sequences (which have ground-truth MANO) as our test split.

\paragraph{HOI4D~\cite{hoi4d_cvpr22}.} A 4D egocentric dataset for category-level human--object interaction, captured at $15$\,fps---half the $30$\,fps used by the other datasets in our suite. It covers 800+ object instances across 16 categories with rich contact and occlusion patterns. We use HOI4D \emph{exclusively as a held-out test set}, and none of the eight baselines is trained on HOI4D either. Most clips depict \emph{single-handed} manipulation, so the second slot must report ``no hand on screen'' for hundreds of consecutive frames without predicting a spurious second hand; several baselines show substantially lower left-hand detection F1 on HOI4D as a result (Tab.~\ref{tab:detection-summary}). The active hand also frequently enters and exits through the frame boundary mid-clip, so the visibility flag must flip at the correct frames rather than being set once per sequence. The lower frame rate enlarges inter-frame motion, making temporal modeling more challenging.

\begin{table}[!htbp]
\caption{\textbf{Dataset specifications.} Train and test segment counts, frame rate, total duration, image resolution after the height-to-$480$ preprocessing, post-resize pinhole intrinsics ($f_x$, $f_y$, $c_x$, $c_y$), horizontal field of view, the supervised overlay target during VACE training (joint skeleton or full mesh; ``--'' for held-out), and the dominant property that motivates including each dataset. EgoDex is the Stage-1a pretraining corpus; ARCTIC and HOT3D supply both the Stage-1b finetuning data and the Stage-2 decoder training data; HOI4D is used only as an out-of-distribution test set, so it has no training segments. Each video is decomposed into non-overlapping $81$-frame segments, the unit on which all reported counts and metrics are computed; any final tail shorter than $81$ frames is excluded so that all segments share the same temporal horizon.}
\label{tab:dataset-spec-supp}
\centering
\resizebox{\textwidth}{!}{%
\begin{tabular}{l cc cc ccccc c cl}
\toprule
Dataset & \# train & \# test & FPS & Hours & $W{\times}H$ & $f_x$ & $f_y$ & $c_x$ & $c_y$ & HFOV & Overlay & Key property \\
\midrule
EgoDex~\cite{egodex}        & 913{,}559 & --  & 30 & 685.2 & varies              & varies & varies & varies & varies & varies              & joint      & large-scale, diverse scenes \\
ARCTIC~\cite{arctic_cvpr23} & 2{,}145   & 291 & 30 & 1.83  & $672{\times}480$    & 579.4  & 579.2  & 318.9  & 235.9  & ${\sim}60^\circ$   & mesh       & bimanual; severe hand--obj./hand--hand occlusion \\
HOT3D~\cite{hot3d_cvpr25}   & 5{,}432   & 437 & 30 & 4.40  & $480{\times}480$    & 207.8  & 207.8  & 239.8  & 239.8  & ${\sim}99^\circ$   & mesh       & fisheye; HDR lighting; large head/hand motion blur \\
HOI4D~\cite{hoi4d_cvpr22}   & --        & 498 & 15 & 0.75  & $854{\times}480$    & 471.6  & 471.8  & 432.1  & 232.6  & ${\sim}85^\circ$   & --         & out-of-distribution test set; predominantly single-handed \\
\bottomrule
\end{tabular}%
}
\end{table}

\paragraph{Baseline training exposure on ARCTIC and HOT3D.}
For an auditable in-distribution comparison, Tab.~\ref{tab:baseline-exposure} records which methods are themselves supervised on ARCTIC or HOT3D---counting supervised training or fine-tuning of the released checkpoint or a key component (backbone, motion prior, infiller), but not evaluation-only use. The ARCTIC and HOT3D blocks are thus \emph{not} \nickname{}-in-domain against uniformly out-of-domain baselines: WiLoR and HaWoR train on both, WildHands on ARCTIC, and Dyn-HaMR inherits ARCTIC through its motion prior, so \nickname{}'s lead there is measured against several \emph{in-domain} competitors, while for InterWild, HaMeR, Hamba, and the default OmniHands it also reflects a distribution advantage. Two caveats follow. The OmniHands figure depends on the checkpoint; we use the released default, which excludes ARCTIC. More sensitively, because HOT3D publishes no ground truth for its official test set, our HOT3D split is $5\%$ of the validation sequences (\S\ref{sec:supp-implementation}); if WiLoR's or HaWoR's HOT3D training overlaps those specific sequences, their HOT3D numbers could be optimistic, and a fully auditable comparison would match sequence IDs. We report this as a reporting/auditability limitation rather than evidence of unfair training; it does not affect HOI4D, which is held out from every method (\S\ref{sec:supp-tponly}).

\begin{table}[!htbp]
\caption{\textbf{Baseline training exposure on the in-distribution benchmarks.} Whether each method's released default checkpoint---or a key component---is supervised on ARCTIC/HOT3D, per its published training recipe; evaluation-only use is excluded. \cmark: trained; \xmark: not trained; (ind.): indirect exposure through a pretrained component. \nickname{} finetunes on an ARCTIC+HOT3D mixture (Stage~1b); no method, ours included, trains on HOI4D.}
\label{tab:baseline-exposure}
\centering
\small
\begin{tabular}{@{}l c c >{\raggedright\arraybackslash}p{7.0cm}@{}}
\toprule
Method & ARCTIC & HOT3D & Training data (per original paper) \\
\midrule
InterWild~\cite{interwild_cvpr23}  & \xmark & \xmark & InterHand2.6M, MSCOCO \\
HaMeR~\cite{hamer_cvpr24}          & \xmark & \xmark & FreiHAND, HO3D, InterHand2.6M, DexYCB, COCO, etc. \\
Hamba~\cite{hamba_nips24}          & \xmark & \xmark & HaMeR recipe ($2.7$M samples) \\
WildHands~\cite{wildhands_eccv24}  & \cmark & \xmark & ARCTIC, AssemblyHands, EPIC; init.\ ArcticNet-SF \\
OmniHands~\cite{omnihands_arxiv24} & \xmark$^\dagger$ & \xmark & InterHand2.6M, Re:InterHand, DexYCB \\
WiLoR~\cite{wilor_cvpr25}          & \cmark & \cmark & regressor training set includes ARCTIC and HOT3D \\
Dyn-HaMR~\cite{dynhamr_cvpr25}     & (ind.) & \xmark & HMP motion prior pretrained on ARCTIC \\
HaWoR~\cite{hawor_cvpr25}          & \cmark & \cmark & HOT3D ($573$K)\,+\,ARCTIC ($165$K) frames; infiller on HOT3D \\
\midrule
\nickname{} (Ours)                 & \cmark & \cmark & Stage-1b mesh-overlay finetuning (ARCTIC+HOT3D) \\
\bottomrule
\end{tabular}
\\[3pt]
{\footnotesize $^\dagger$Default checkpoint excludes ARCTIC; an ``I+R+A'' variant adds it, so the exact checkpoint matters.}
\end{table}

\subsection{Decoder Architecture}
\label{sec:supp-projector}

The decoder is a $37$-million-parameter network that decodes the fixed intermediate DiT activations of the Stage-1b backbone into all MANO parameters in a single forward pass. It comprises four modules that match the main paper: a Hand-Token Branch that produces per-hand parametric tokens via cross-attention with ray-space positional encoding; a parallel Joint-Heatmap Branch that produces $21$ per-hand 2D joint estimates and per-joint visual descriptors; a Hand--Joint Fusion layer in which the two streams refine each other; and a Mixed-Projection Head that regresses pose, shape, depth, and on-screen visibility while solving the in-plane translation in closed form against the heatmap. Every multi-head attention layer uses eight heads. For brevity we drop the slot $s$ and time $t$ indices throughout this section; the unsubscripted symbols $\hat R, \widehat{\boldsymbol\theta}, \widehat{\boldsymbol\beta}, \hat\zeta, \hat e, \widehat{\mathbf{p}}^{\mathrm{init}}, \widehat{\mathbf{p}}^{\mathrm{final}}$ denote per-slot, per-frame entries of the slot-stacked bold symbols $\widehat{\mathbf{R}}_t, \widehat{\boldsymbol\Theta}_t, \widehat{\mathbf{B}}_t, \widehat{\boldsymbol\zeta}_t, \widehat{\mathbf{e}}_t, \widehat{\mathbf{P}}^{\mathrm{init}}_t, \widehat{\mathbf{P}}^{\mathrm{final}}_t$ of the main paper.

\paragraph{Hand-Token Branch.} The $1536$-channel DiT feature tensor is reduced to $512$ channels by a linear projection followed by layer normalization, and three additive positional encodings are applied: a learned spatial encoding, a learned temporal encoding, and a ray-space positional encoding (\S\ref{sec:supp-ray-pe}). Two learned hand queries with deliberately different initial offsets cross-attend to all $H{\cdot}W$ spatial tokens through four shared transformer-decoder layers, each consisting of self-attention, dense cross-attention, and a feed-forward network with hidden dimension $2048$ and GELU activation. The output is $Q^{\mathrm{hand}} \in \mathbb{R}^{2{\times}512}$, one query per hand slot $s\in\{\mathrm{L},\mathrm{R}\}$. Because handedness is assigned by slot, no handedness classifier or query-division block is required.

\paragraph{Joint-Heatmap Branch.} In parallel, a $1{\times}1$ convolution applied to the spatial features produces, per slot $s\in\{\mathrm{L},\mathrm{R}\}$, $21$ joint heatmaps $\mathcal{H}_s \in \mathbb{R}^{J{\times}H{\times}W}$ with $J{=}21$. A differentiable soft-argmax over each heatmap yields an initial 2D estimate $\widehat{\mathbf{p}}^{\mathrm{init}}_s \in \mathbb{R}^{J{\times}2}$ in normalized $[0,1]$ image coordinates. Pooling the spatial features with $\mathrm{softmax}(\mathcal{H}_s)$ along the spatial axis produces $21$ joint-level features $Q^{\mathrm{joint}}_s \in \mathbb{R}^{J{\times}512}$ that summarize the visual evidence at each predicted joint location of slot $s$.

\paragraph{Hand--Joint Fusion.} A single mutual cross-attention layer fuses the two streams, with eight attention heads, residual connections, and post-layer-normalization:
\begin{align}
\widetilde{Q}^{\mathrm{hand}} &= \mathrm{LN}\!\big(Q^{\mathrm{hand}} + \mathrm{MHA}(Q^{\mathrm{hand}}, Q^{\mathrm{joint}}, Q^{\mathrm{joint}})\big), \\
\widetilde{Q}^{\mathrm{joint}} &= \mathrm{LN}\!\big(Q^{\mathrm{joint}} + \mathrm{MHA}(Q^{\mathrm{joint}}, Q^{\mathrm{hand}}, Q^{\mathrm{hand}})\big),
\end{align}
where $\mathrm{MHA}(\cdot,\cdot,\cdot)$ denotes a multi-head attention call with the three arguments serving as queries, keys, and values, and $\mathrm{LN}$ denotes layer normalization. The output projection of each multi-head attention block and the last layer of the offset network below are zero-initialized; at initialization the cross-attention residual is exactly zero and the Joint-Heatmap Branch passes its initial 2D estimates through unchanged, so training begins from a state in which the two streams are independent and only the fusion residual has to be learned. An offset network fed by $\widetilde{Q}^{\mathrm{joint}}$ produces per-joint 2D offsets $\Delta\mathbf{p}$, and the refined heatmap-derived 2D joints are
\begin{equation}
\widehat{\mathbf{p}}^{\mathrm{final}} = \widehat{\mathbf{p}}^{\mathrm{init}} + \Delta\mathbf{p}.
\end{equation}

\paragraph{Mixed-Projection Head.} The fused hand features $\widetilde{Q}^{\mathrm{hand}}$ feed two branches, one for visibility and one for regression.

The on-screen visibility branch consists of one decoder layer followed by a two-layer MLP of hidden dimension $128$, and is gradient-detached from the shared features so that the dominant regression gradients do not overwhelm the binary visibility signal. It produces a single on-screen visibility logit $\hat e$ per slot and is upsampled to video frames by linear interpolation; no handedness output is produced because handedness is fixed by slot.

The regression branch consists of two decoder layers that receive full gradient from the shared features, and predicts the global orientation $\hat R \in \mathrm{SO}(3)$, the $15$ finger-joint rotations $\widehat{\boldsymbol\theta} \in \mathrm{SO}(3)^{15}$, the $10$ shape coefficients $\widehat{\boldsymbol\beta} \in \mathbb{R}^{10}$, and the log-depth scalar $\hat\zeta \in \mathbb{R}$, each via a two-layer MLP with hidden dimension $256$. The in-plane translation $(\hat t^x, \hat t^y)$ is \emph{not} produced by this head; instead it is solved analytically (\S\ref{sec:supp-mixed-proj}) from the refined heatmap anchors $\widehat{\mathbf{p}}^{\mathrm{final}}$, the regressed depth $\hat t^z = \exp(\hat\zeta)$, and the camera intrinsics. Because the cached features live at the latent frame rate ($T_{\mathrm{lat}}{=}21$) while MANO supervision lives at the video frame rate ($T{=}81$), the regression branch upsamples its latent-frame outputs to video frames before loss computation. The upsampler proceeds in three stages: linear interpolation lifts the signal to video frame rate; a two-layer 1D convolution with kernel size~$5$ and GELU activation provides a residual refinement with a zero-initialized output; and a depthwise-separable 1D convolution with kernel size~$7$ acts as a learned temporal mixer for smoothing. Rotations are produced in the 6D continuous rotation representation and converted to $3{\times}3$ matrices via Gram--Schmidt orthogonalization.

\subsubsection{Ray-Space Positional Encoding}
\label{sec:supp-ray-pe}

For each spatial token at grid position $(i, j)$, we compute the camera ray direction via the pinhole camera model:
\begin{align}
u_\mathrm{px} &= (j + 0.5) \cdot W_\mathrm{img} / W_\mathrm{pat}, \quad v_\mathrm{px} = (i + 0.5) \cdot H_\mathrm{img} / H_\mathrm{pat}, \\
\alpha &= \arctan\!\Big(\frac{u_\mathrm{px} - c_x}{f_x}\Big), \quad \eta = \arctan\!\Big(\frac{v_\mathrm{px} - c_y}{f_y}\Big),
\end{align}
where $(f_x, f_y, c_x, c_y)$ are the camera intrinsics. The azimuth and elevation are encoded with eight sinusoidal frequency bands,
\begin{equation}
\mathbf{e}_{ij} = \big[\sin(\omega\alpha),\, \cos(\omega\alpha),\, \sin(\omega\eta),\, \cos(\omega\eta)\big]_{\omega\in\Omega} \in \mathbb{R}^{32},\quad \Omega = \{2^0, 2^1, \ldots, 2^7\},
\end{equation}
projected through a two-layer multi-layer perceptron with hidden dimension $512$, GELU activation, and a zero-initialized output, and added residually to the learned spatial positional encoding. The zero initialization ensures the ray-space contribution is exactly zero at the start, so the positional encoding initially equals the spatial encoding alone and the network gradually learns camera-aware corrections. Because the encoding is computed per token and per sample, different intrinsics within a batch produce different positional encodings, enabling the cross-attention to adapt to the camera field of view.

\subsubsection{Mixed-Projection Camera-Translation Head}
\label{sec:supp-mixed-proj}

The translation $\mathbf{t} = (t^x, t^y, t^z)$ is recovered by a hybrid scheme rather than fully regressed. The regression branch predicts only the log-depth scalar $\hat\zeta$, from which $\hat t^z = \exp(\hat\zeta)$. Given the regressed $\hat R$, $\widehat{\boldsymbol\theta}$, $\widehat{\boldsymbol\beta}$, the differentiable MANO layer $\mathcal{M}_s(\hat R, \widehat{\boldsymbol\theta}, \widehat{\boldsymbol\beta})$ produces $J{=}21$ root-relative 3D joints $\{(\hat X_j, \hat Y_j, \hat Z_j)\}_{j=0}^{J-1}$ per hand, with $j{=}0$ the wrist (consistent with the metric-section convention of \S\ref{sec:supp-metrics}). For each joint $j$, the pinhole projection
\begin{align}
u_j = f_x\,\frac{\hat X_j + t^x}{\hat Z_j + \hat t^z} + c_x,\qquad
v_j = f_y\,\frac{\hat Y_j + t^y}{\hat Z_j + \hat t^z} + c_y
\end{align}
is linear in $(t^x, t^y)$ once $\hat t^z$ is fixed. Crucially, $u_j$ depends only on $t^x$ and $v_j$ only on $t^y$, so the 2D solve decouples into two independent scalar least-squares problems---equivalent to a $2{\times}2$ system whose matrix is exactly diagonal. Let $(\hat u_j, \hat v_j) = (W\,\hat p^{\mathrm{final},x}_j, H\,\hat p^{\mathrm{final},y}_j)$ denote the refined heatmap anchors converted from normalized $[0,1]$ to pixel coordinates, $z_j = \hat Z_j + \hat t^z$, and $M^{\mathrm{dir}}_j \in \{0,1\}$ a per-joint validity mask using the same heatmap-representability gate as the direct 2D loss $\mathcal{L}_{\mathrm{2D}}$ (\S\ref{sec:supp-losses}). The closed-form solution is
\begin{equation}
\hat t^x =
\frac{\sum_{j} M^{\mathrm{dir}}_j\,\frac{f_x}{z_j}\!\left(\hat u_j - c_x - \frac{f_x \hat X_j}{z_j}\right)}
{\sum_{j} M^{\mathrm{dir}}_j \!\left(\frac{f_x}{z_j}\right)^{\!2}},\qquad
\hat t^y =
\frac{\sum_{j} M^{\mathrm{dir}}_j\,\frac{f_y}{z_j}\!\left(\hat v_j - c_y - \frac{f_y \hat Y_j}{z_j}\right)}
{\sum_{j} M^{\mathrm{dir}}_j \!\left(\frac{f_y}{z_j}\right)^{\!2}}.
\label{eq:supp-mixed-proj}
\end{equation}
All $21$ joints participate when their anchor is on-screen and inside the heatmap-representable range; the remaining joints are masked out, so the effective system size adapts per frame and per hand rather than being fixed at a hand-crafted subset. This gate is evaluated on the \emph{ground-truth} anchor during training (matching $\mathcal{L}_{\mathrm{2D}}$), but \emph{at inference no ground truth is available or used}: a joint enters the solve when its \emph{predicted} refined-heatmap anchor $\widehat{\mathbf{p}}^{\mathrm{final}}$ lies in the image and its regressed depth places it in front of the camera ($\hat z_j > 0.05$\,m), so the reported camera translation is computed entirely from the model's own predictions.

This decomposes the camera translation $\mathbf{t}$ into a regressed depth, where image evidence is least informative and a scalar prediction is appropriate, and a closed-form in-plane translation, where the Joint-Heatmap Branch already provides pixel-level evidence and the analytic solve removes one ill-conditioned regression target. The solve is differentiable, so gradients from $\mathcal{L}_{\mathrm{trans}}$ propagate back into the regression head (via $\hat X_j, \hat Y_j, \hat Z_j, \hat\zeta$) and the heatmap branch (via $\widehat{\mathbf{p}}^{\mathrm{final}}$), tying MANO pose, depth, and 2D anchoring into one coupled geometric system. Ray-space PE and the mixed-projection solve are complementary: the former adapts the cross-attention features to the camera field of view at the perceptual stage, while the latter converts 2D predictions to camera-frame 3D analytically using the camera intrinsics.

\subsection{Loss Functions}
\label{sec:supp-losses}

The total decoder loss $\mathcal{L}_{\mathrm{dec}}$ has the grouped form introduced in the main paper; Tab.~\ref{tab:loss-details} expands that grouped expression into the ten individually weighted terms, and the per-term paragraphs below give the design rationale of each.

\begin{table}[!htbp]
\caption{\textbf{Loss terms and weights.} Mathematical form, weight $\lambda$, and short description of each of the ten decoder loss terms; the full mathematical definitions and design rationale follow in the paragraphs below. All terms are computed per frame and per hand and gated by four binary validity masks: $M_{t,s}$ (frame $t$ and slot $s$ have a valid ground-truth on screen), $M^{\mathrm{proj}}_{t,s,j}$ (the ground-truth projection of joint $j$ falls inside the image plane), $M^{\mathrm{dir}}_{t,s,j}$ (the ground-truth 2D position of joint $j$ falls inside the heatmap-representable range), and $M^{\mathrm{tri}}_{t,s}$ (frames $t{-}1$, $t$, and $t{+}1$ all have valid predictions for slot $s$). The largest weight is on the assembled camera translation ($\lambda_{\mathrm{trans}}{=}3.0$) and the smallest on shape consistency ($\lambda_{\mathrm{shape}}^{\mathrm{con}}{=}0.05$).}
\label{tab:loss-details}
\centering
\small
\begin{tabular}{llcl}
\toprule
Term & Form & $\lambda$ & Description \\
\midrule
$\mathcal{L}_{\mathrm{orient}}$              & $d_{\mathrm{SO}(3)}(\hat R, R^\star)$                                                       & 0.2  & Global orientation (geodesic) \\
$\mathcal{L}_{\mathrm{pose}}$                & $\tfrac{1}{15}\sum_{k} d_{\mathrm{SO}(3)}(\hat\theta_k, \theta^\star_k)$                    & 0.2  & Finger joint rotations (geodesic) \\
$\mathcal{L}_{\mathrm{shape}}$               & $\|\widehat{\boldsymbol\beta} - \boldsymbol\beta^\star\|_2^2$                                   & 0.1  & Shape coefficients \\
$\mathcal{L}_{\mathrm{trans}}$               & $\|\widehat{\mathbf{t}} - \mathbf{t}^\star\|_2^2$                                               & 3.0  & Camera translation (mixed-projection output) \\
$\mathcal{L}_{\mathrm{3D}}$                  & $\|\widehat{\mathbf{J}} - \mathbf{J}^\star\|_1$                                                 & 2.0  & 3D joints (camera frame) \\
$\mathcal{L}_{\mathrm{reproj}}$              & $\|\pi_K(\widehat{\mathbf{J}}) - \mathbf{p}^\star\|_1$                                          & 1.0  & 2D reprojection (from MANO mesh) \\
$\mathcal{L}_{\mathrm{2D}}$                  & $\|\widehat{\mathbf{p}}^{\mathrm{final}} - \mathbf{p}^\star\|_1$                                & 2.0  & Heatmap-derived 21-joint 2D \\
$\mathcal{L}_{\mathrm{vis}}$                 & $\mathrm{BCE}(\hat e, e^\star)$                                                             & 0.5  & On-screen visibility (handedness fixed by slot) \\
$\mathcal{L}_{\mathrm{acc}}$                 & $\|\widehat{\mathbf{t}}_{t+1} - 2\widehat{\mathbf{t}}_t + \widehat{\mathbf{t}}_{t-1}\|_1$               & 0.3  & Translation acceleration smoothness \\
$\mathcal{L}_{\mathrm{shape}}^{\mathrm{con}}$ & $\|\widehat{\boldsymbol\beta} - \mathrm{sg}(\overline{\boldsymbol\beta})\|_2^2$                      & 0.05 & Shape consistency (stop-gradient mean) \\
\bottomrule
\end{tabular}
\end{table}

We describe each loss term and its design rationale below. All rotation predictions are produced in the 6D continuous rotation representation and converted to $3{\times}3$ matrices via Gram--Schmidt orthogonalization before loss computation.

\paragraph{Geodesic rotation losses $\mathcal{L}_{\mathrm{orient}}, \mathcal{L}_{\mathrm{pose}}$.}
We supervise global orientation $\hat R \in \mathrm{SO}(3)$ ($\mathcal{L}_{\mathrm{orient}}$) and $15$ finger joint rotations $\widehat{\boldsymbol\theta} \in \mathrm{SO}(3)^{15}$ ($\mathcal{L}_{\mathrm{pose}}$) with the geodesic distance on the rotation manifold,
\begin{equation}
d_{\mathrm{SO}(3)}(\hat R, R^\star) = \arccos\!\bigl((\mathrm{tr}(\hat R^\top R^\star) - 1)\,/\,2\bigr).
\end{equation}
The geodesic distance is the unique bi-invariant metric on $\mathrm{SO}(3)$; it measures the \emph{angular} distance between two rotations regardless of axis. Element-wise mean squared error on the $3{\times}3$ matrix, in contrast, treats off-diagonal entries as independent scalars, ignoring the manifold structure and producing gradients that can push predictions off $\mathrm{SO}(3)$. The loss ablation on ARCTIC (Tab.~\ref{tab:loss-ablation-main}) shows that replacing the geodesic supervision with element-wise mean squared error produces the largest single-loss degradation in the table on MPJPE-p ($+0.84$\,mm), consistent with the requirement that articulated hand-pose supervision respect the rotation manifold.

\paragraph{Camera translation loss $\mathcal{L}_{\mathrm{trans}}$.}
The camera translation $\widehat{\mathbf{t}} = (\hat t^x, \hat t^y, \hat t^z) \in \mathbb{R}^3$ assembled by the Mixed-Projection Head (\S\ref{sec:supp-mixed-proj}) is supervised with mean squared error and assigned the highest per-term weight ($\lambda_{\mathrm{trans}}{=}3.0$). The high weight reflects two design considerations. First, translation controls absolute hand position in the camera frame and is central to 3D reconstruction quality. Second, $\hat t^z$ enters the head as the regressed $\hat\zeta$ and $(\hat t^x, \hat t^y)$ are obtained from the closed-form per-coordinate solve, so the raw prediction scale is small relative to the MANO rotation outputs and a higher loss weight is needed to balance gradient magnitudes. The error is applied to the fully assembled $(t^x, t^y, t^z)$, so gradients flow back through the Joint-Heatmap Branch via $\widehat{\mathbf{p}}^{\mathrm{final}}$ as well as through the regression branch's $\hat\zeta$ and 3D-joint outputs.

\paragraph{Shape coefficient loss $\mathcal{L}_{\mathrm{shape}}$.}
The 10-dimensional MANO shape coefficients $\widehat{\boldsymbol\beta}$ are supervised with mean squared error. Hand shape varies slowly and is typically constant within a clip, so a low weight ($\lambda_{\mathrm{shape}}{=}0.1$) is sufficient to anchor the prediction without dominating the joint losses that drive articulated geometry.

\paragraph{On-screen visibility loss $\mathcal{L}_{\mathrm{vis}}$.}
A single binary cross-entropy term $\mathrm{BCE}(\hat e, e^\star)$ supervises whether any of the $21$ ground-truth MANO joints for this slot projects into the image plane with $z > 0.01$\,m, the same on-screen criterion used by the off-screen-exclusion policy at evaluation time (\S\ref{sec:supp-alignment}). Because handedness is fixed by slot ($s\in\{\mathrm{L},\mathrm{R}\}$) and a 3D slot-validity flag that is invisible on screen provides no useful supervision signal, we drop the separate handedness and 3D-presence cross-entropy terms used in earlier transformer-decoder hand pipelines and supervise only the on-screen flag. The loss is computed on all slots, including empty ones, so the model is explicitly trained to suppress predictions when no hand is on screen rather than to silently ignore absent-hand slots. For numerical stability, the per-sample term is clipped at $5.0$.

\paragraph{3D joint L1 loss $\mathcal{L}_{\mathrm{3D}}$.}
We run a differentiable batched MANO forward pass $\mathcal{M}_s(\hat R, \widehat{\boldsymbol\theta}, \widehat{\boldsymbol\beta})$ to obtain $21$ root-relative joints $\widehat{\mathbf{J}}^0$ per hand, add the predicted translation $\widehat{\mathbf{t}}$, and supervise the camera-frame joints with L1,
\begin{equation}
\mathcal{L}_{\mathrm{3D}} = \bigl\|\widehat{\mathbf{J}} - \mathbf{J}^\star\bigr\|_1, \qquad \widehat{\mathbf{J}} = \mathcal{M}_s(\hat R, \widehat{\boldsymbol\theta}, \widehat{\boldsymbol\beta}) + \widehat{\mathbf{t}}.
\end{equation}
An $\ell_1$ loss is preferred over $\ell_2$ for robustness to outlier joints (e.g., occluded fingertips with noisy ground truth). This term provides direct geometric supervision on the final 3D output and carries weight $\lambda_{\mathrm{3D}}{=}2.0$.

\paragraph{2D reprojection L1 loss $\mathcal{L}_{\mathrm{reproj}}$.}
The predicted camera-frame 3D joints $\widehat{\mathbf{J}}$ are projected to normalized image coordinates via the pinhole model with dataset-provided intrinsics $(f_x, f_y, c_x, c_y)$, and supervised against the 2D ground truth $\mathbf{p}^\star$,
\begin{equation}
\mathcal{L}_{\mathrm{reproj}} = \bigl\|\pi_K(\widehat{\mathbf{J}}) \;-\; \mathbf{p}^\star\bigr\|_1, \quad
\mathbf{p}^\star \in [0,1]^{21 \times 2}.
\end{equation}
Both predictions and targets are normalized to $[0,1]$ by dividing by image dimensions, making the loss scale-invariant across the two training datasets, which have different resolutions (ARCTIC $672{\times}480$ and HOT3D $480{\times}480$). Depth values are clamped to $z \geq 0.05$\,m to prevent numerical instability from near-zero denominators. Rather than clamping off-screen projections, we apply a per-joint on-screen mask ($M^{\mathrm{proj}}_{t,s,j}$ in the main paper) computed from the ground-truth 2D projection: only joints whose ground-truth position falls within $[0, W) \times [0, H)$ with $z > 0.01$\,m receive 2D supervision, and the loss is normalized by the total count of visible joints across the batch. This unbiased masking avoids penalizing correct 3D predictions that happen to project off-screen. The 2D reprojection term resolves the depth--rotation gauge ambiguity, in which multiple combinations of depth and wrist rotation can produce the same 3D joint constellation in camera frame but only the correct depth yields accurate 2D projections. By grounding supervision in pixel space, the loss provides an implicit intrinsics-aware signal for depth-sensitive reconstruction; it is among the most impactful entries of the loss ablation in Tab.~\ref{tab:loss-ablation-main}.

\paragraph{Heatmap-2D loss $\mathcal{L}_{\mathrm{2D}}$.}
The heatmap branch produces refined per-joint 2D coordinates $\widehat{\mathbf{p}}^{\mathrm{final}} = \widehat{\mathbf{p}}^{\mathrm{init}} + \Delta\mathbf{p}$ in normalized $[0,1]$ image coordinates (\S\ref{sec:supp-projector}). We supervise these directly with an $\ell_1$ loss against the ground-truth 2D joint positions $\mathbf{p}^\star$,
\begin{equation}
\mathcal{L}_{\mathrm{2D}} = \bigl\|\widehat{\mathbf{p}}^{\mathrm{final}} - \mathbf{p}^\star\bigr\|_1, \quad
\mathbf{p}^\star \in [0,1]^{21 \times 2}.
\end{equation}
This term anchors the Joint-Heatmap Branch independently of the MANO mesh: $\widehat{\mathbf{p}}^{\mathrm{final}}$ flows into the Mixed-Projection Head as the observation in the closed-form $(t^x, t^y)$ solve (\S\ref{sec:supp-mixed-proj}), so a clean heatmap is a prerequisite for clean in-plane translation. The direct-2D mask $M^{\mathrm{dir}}_{t,s,j}$ (heatmap-representable joints) gates this term. Without it, $\widehat{\mathbf{p}}^{\mathrm{final}}$ is supervised only indirectly through the MANO mesh, the heatmap converges much later, and the camera translation is poorly anchored in early training.

\paragraph{Translation acceleration smoothness $\mathcal{L}_{\mathrm{acc}}$.}
We penalize the second-order finite difference (discrete acceleration) of predicted camera translations, masked to frame triples in which all three frames have a valid prediction,
\begin{equation}
\mathcal{L}_{\mathrm{acc}} = \frac{\sum_{s\in\{\mathrm{L},\mathrm{R}\}}\sum_{t=2}^{T-1} M^{\mathrm{tri}}_{t,s}\,\bigl\|\widehat{\mathbf{t}}_{t+1,s} - 2\widehat{\mathbf{t}}_{t,s} + \widehat{\mathbf{t}}_{t-1,s}\bigr\|_1}{\max\bigl(\sum_{s\in\{\mathrm{L},\mathrm{R}\}}\sum_{t=2}^{T-1} M^{\mathrm{tri}}_{t,s},\;1\bigr)},
\end{equation}
where $M^{\mathrm{tri}}_{t,s} \in \{0,1\}$ is the triple-frame validity mask defined in the caption of Tab.~\ref{tab:loss-details}. The second-order finite difference penalizes abrupt acceleration changes (jitter) while allowing smooth velocity changes, which is more appropriate for natural hand motion than a first-order velocity penalty that would resist all motion. The $\ell_1$ form provides robustness to occasional large accelerations during fast hand movements. The mask avoids spurious penalties across temporal gaps. Removing this term substantially degrades the Jitter metric (Tab.~\ref{tab:loss-ablation-main}).

\paragraph{Shape consistency loss $\mathcal{L}_{\mathrm{shape}}^{\mathrm{con}}$.}
This term regularizes every per-frame shape prediction toward the per-slot temporal mean of the predictions within the clip,
\begin{equation}
\mathcal{L}_{\mathrm{shape}}^{\mathrm{con}} = \frac{1}{\sum_s |\mathcal{V}_s|} \sum_{s\in\{\mathrm{L},\mathrm{R}\}} \sum_{t\in\mathcal{V}_s} \|\widehat{\boldsymbol\beta}_{t,s} - \mathrm{sg}(\overline{\boldsymbol\beta}_s)\|_2^2, \qquad \overline{\boldsymbol\beta}_s = \frac{1}{|\mathcal{V}_s|}\sum_{t \in \mathcal{V}_s} \widehat{\boldsymbol\beta}_{t,s},
\end{equation}
where $\mathcal{V}_s$ is the set of valid frames for slot $s$ and $\mathrm{sg}(\cdot)$ is the stop-gradient operator. Both the inner sum and the mean are taken per slot, so the left and right hand shape coefficients regularize independently. The temporal mean $\overline{\boldsymbol\beta}_s$ is detached from the computation graph: gradients flow only through $\widehat{\boldsymbol\beta}_{t,s}$, not through $\overline{\boldsymbol\beta}_s$. The stop-gradient prevents mode collapse, since without it the loss could be minimized by collapsing all per-frame predictions to a single degenerate point. Individual frames are instead pulled toward their current average, enforcing the physical prior that hand shape does not change within a short clip. The low weight ($\lambda_{\mathrm{shape}}^{\mathrm{con}}{=}0.05$) reflects that this term is a soft regularizer rather than a primary supervision signal.

\subsection{Two-Stage Training Pipeline}
\label{sec:supp-training}

Stage~1 adapts the VACE branch in two sub-steps (Stage~1a and Stage~1b), and Stage~2 trains the MANO decoder on cached features from the Stage~1b backbone.

\subsubsection{Stage 1a: Joint-Overlay Pretraining on EgoDex}

Stage~1a pretrains the VACE conditioning path on EgoDex~\cite{egodex} alone, a large-scale egocentric hand--object manipulation corpus with 3D joint annotations but no MANO mesh labels; no MANO-annotated dataset is mixed in at this sub-step. The rendered target is a joint-skeleton overlay alpha-blended with the scene, optimized for $25$\,k steps with AdamW at learning rate $10^{-4}$ and a cosine schedule with a $500$-step linear warmup. The run uses $32$ GPUs at per-GPU batch size one in bfloat16 mixed precision with DeepSpeed ZeRO stage~$0$.

\subsubsection{Stage 1b: MANO Mesh-Overlay Finetuning}

Starting from the Stage~1a checkpoint, we continue training for $10$\,k steps on the two in-distribution MANO-annotated egocentric datasets (ARCTIC, HOT3D) with segment-proportional sampling weights ($0.283$, $0.717$ respectively) and the full mesh-overlay target. HOI4D is held out from every explicit supervised stage of our pipeline (Stage~1a, Stage~1b, and decoder training), so the main paper's HOI4D evaluation is a cross-dataset test with respect to our supervised training, on the same footing as the baselines, none of which is finetuned on HOI4D either; as with any pretrained video foundation model, we cannot audit whether HOI4D-like footage appears in Wan2.1-VACE's own pretraining corpus (see the main paper's Limitations). Stage~1b runs on $8$ GPUs with the same per-GPU batch size, optimizer, and schedule as Stage~1a. As a control, we also train Stage~1b from scratch without EgoDex initialization for $25$\,k steps on $8$ GPUs, matching the Stage-1a step budget; this configuration is the \emph{Mesh-overlay only} row of the main paper's data-ablation table.

\subsubsection{Stage 2: MANO Decoder Training}

The decoder is trained on a fixed feature slice extracted from the Stage~1b VACE backbone. For each segment, we run one $50$-step sampling pass (schedule shift $5$) and record the DiT $L_{15}$ activations at step $34$, i.e.\ normalized denoising step $\tau_k \approx 0.7$, yielding a feature tensor of shape $[1536, F_{\mathrm{lat}}, H_{\mathrm{pat}}, W_{\mathrm{pat}}]$ per segment with $F_{\mathrm{lat}}{=}21$ and $(H_{\mathrm{pat}}, W_{\mathrm{pat}})$ as in \S\ref{sec:supp-datasets}. Because the shifted schedule is non-uniform, this step still retains most of the initial noise despite lying about two-thirds of the way along the trajectory, consistent with the main-paper analysis. The decoder reads this slice and is optimized for $30$\,k steps with batch size $16$ using AdamW at learning rate $2{\times}10^{-4}$ and a cosine schedule with a $200$-step linear warmup; each batch is drawn from a single dataset so that per-batch spatial grids match.

\subsection{Controlled Fitting Study: Capacity of the Feature Slice}
\label{sec:supp-fitting}

Before making generalization claims, we verify that the chosen mid-denoise feature slice contains enough hand information to support MANO decoding---if it did not, even overfitting a single sequence would fail to recover accurate and temporally consistent hand motion. With the Stage~1b VACE backbone frozen, we fit the dual-branch decoder of \S\ref{sec:supp-projector} on a single ARCTIC segment using the default decoding point ($L_{15}$, $\tau_k{\approx}0.7$) and the full decoder loss of \S\ref{sec:supp-losses}. After $3$\,k steps the decoder reaches MPJPE-p $0.35$\,mm and PA-MPJPE-p $0.12$\,mm on the held-in clip, well below any per-frame ambiguity that the MANO surface itself supports. This indicates that the selected feature slice contains sufficient information to decode the hand state in a held-in setting; the test-time errors reported in the comparison and ablation tables of the main paper therefore primarily reflect generalization rather than an obvious feature-capacity bottleneck.

\section{Loss-Term Ablation on ARCTIC}
\label{sec:supp-ablations}

This section reports a loss-term ablation on ARCTIC; the decoder-component ablation is given in the main paper. Following the ablation protocol of the main paper, only the Stage-2 decoder training is restricted to ARCTIC alone, while the Stage-1b VACE backbone is the same ARCTIC + HOT3D one used by the main-paper comparison; this isolates the decoder ablation from cross-dataset transfer while holding the backbone fixed. Tab.~\ref{tab:loss-ablation-main} removes one loss term at a time from the full design. The four reported metrics are FAcc, MPJPE-p, EPE-p, and Jitter; FAcc is higher-better and the rest are lower-better.

\begin{table}[H]
\centering
\caption{\textbf{Loss-term ablation on ARCTIC.} Each row removes a single loss term from the full decoder objective, holding all other terms and the training protocol fixed. FAcc is higher-better; the rest are lower-better.}
\label{tab:loss-ablation-main}
\resizebox{0.55\textwidth}{!}{%
\begin{tabular}{lcccc}
\toprule
Variant & FAcc & MPJPE-p & EPE-p & Jitter \\
\midrule
w/o geodesic       & 0.9970 & 21.43 & 11.92 & 3.49 \\
w/o 3D joint L1    & 0.9975 & 21.15 & 12.25 & \textbf{3.39} \\
w/o 2D reproj      & 0.9507 & 21.19 & 11.98 & 3.42 \\
w/o accel.\ smooth & 0.9479 & 21.10 & \textbf{11.84} & 3.88 \\
w/o shape cons.    & \textbf{0.9979} & 20.71 & 12.01 & 3.50 \\
\midrule
Full design        & \textbf{0.9979} & \textbf{20.59} & 11.93 & 3.42 \\
\bottomrule
\end{tabular}%
}
\end{table}

\paragraph{Loss-term ablation.}
Removing any single loss term increases MPJPE-p relative to the full objective, with the shape-consistency loss producing the smallest increase at $0.12$\,mm and the geodesic rotation loss the largest at $0.84$\,mm. Two terms drive specific failure modes on the other metrics. Acceleration smoothness is the only loss that materially changes Jitter, and removing it raises Jitter from $3.42$ to $3.88$\,mm/frame$^2$, confirming its role as a temporal regularizer. Removing the 2D reprojection loss drops FAcc to $0.9507$ and removing the acceleration smoothness loss drops it to $0.9479$, the only two ablations that substantially affect FAcc; we attribute this sensitivity to its strict per-frame criterion, under which a few outlier frames suffice to flip the metric. EPE-p is comparatively flat across the loss-term ablation because the heatmap-2D and 2D-reprojection losses each independently anchor pixel-space accuracy.

\end{document}